\documentclass[10pt,twocolumn,letterpaper]{article}

\usepackage[pagenumbers]{cvpr} 

\definecolor{cvprblue}{rgb}{0.21,0.49,0.74}
\usepackage[pagebackref,breaklinks,colorlinks,allcolors=cvprblue]{hyperref}

\usepackage{hyperref}
\usepackage{url}

\usepackage{graphicx}
\usepackage{amsmath,bm}
\usepackage{amssymb}
\usepackage{booktabs}
\usepackage{enumitem}
\usepackage{multirow}
\usepackage{wrapfig}
\usepackage{makecell}
\usepackage[accsupp]{axessibility}
\usepackage{footnote}
\usepackage{comment}

\usepackage{algorithm}
\usepackage{arydshln} 
\usepackage[
  separate-uncertainty = true,
  multi-part-units = repeat
]{siunitx}

\usepackage{algorithm}
\usepackage[noend]{algorithmic}

\usepackage[capitalize]{cleveref}
\crefname{section}{Sec.}{Secs.}
\Crefname{section}{Section}{Sections}
\Crefname{table}{Table}{Tables}
\crefname{table}{Tab.}{Tabs.}
\def\paperID{} 
\def\confName{CVPR}
\def\confYear{2026}

\renewcommand{\ie}{\textit{i}.\textit{e}.}
\renewcommand{\eg}{\textit{e}.\textit{g}.}

\newcommand{\nickname}{EvObj}
\title{\nickname{}: Learning Evolving Object-centric Representations for 3D Instance Segmentation without Scene Supervision}

\begin{document}
\author{Jiahao Chen \textsuperscript{1,2} \quad  Zihui Zhang \textsuperscript{1,2} \footnotemark[1] \quad  Yafei Yang \textsuperscript{1,2} \quad  Jinxi Li \textsuperscript{1,2} \\ 
\quad Shenxing Wei \textsuperscript{1,2} \quad Zhixuan Sun \textsuperscript{1,2} \quad Bo Yang \textsuperscript{1,2} \\
 \textsuperscript{1} Shenzhen Research Institute, The Hong Kong Polytechnic University \\
 \textsuperscript{2} vLAR Group, The Hong Kong Polytechnic University\\
{\tt\small \{polyjiahao.chen, zihui.zhang\}@connect.polyu.hk, bo.yang@polyu.edu.hk}}


\twocolumn[{%
\renewcommand\twocolumn[1][]{#1}%
    \maketitle
    \begin{center}
        \vspace{-20pt}
        \centering
        \includegraphics[width=1.0\linewidth]{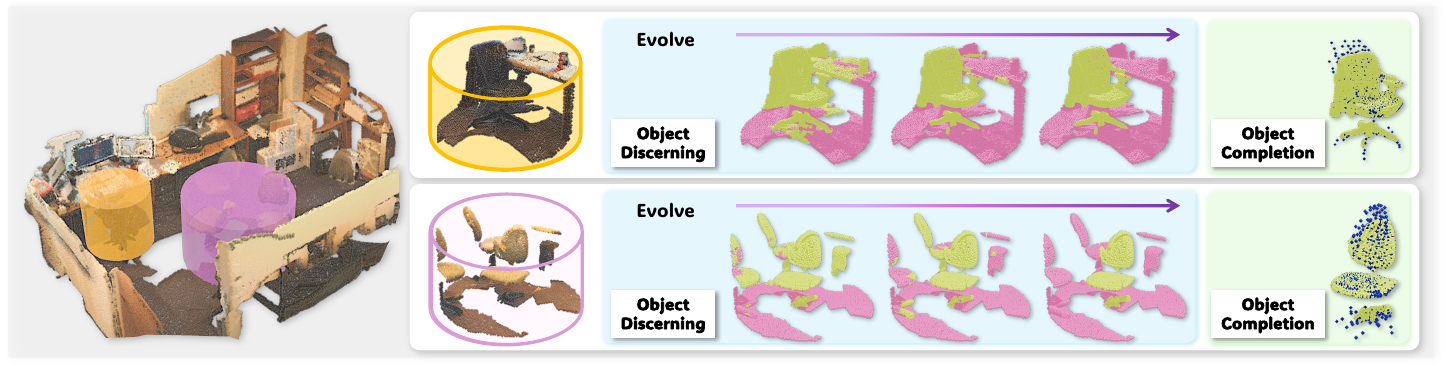}
        \vspace{-20pt}
        \captionof{figure}{\textcolor{black}{Given 3D scene point clouds, our \nickname{} effectively segments complex objects by discerning, refining, and completing object candidates, without needing human annotations in training.}}
        \label{fig:opening}
        \vspace{0pt}
    \end{center}
}]

\renewcommand{\thefootnote}{\fnsymbol{footnote}}
\footnotetext[1]{ Corresponding author}
\maketitle

\begin{abstract}
We introduce \nickname{} for unsupervised 3D instance segmentation that bridges the geometric domain gap between synthetic pretraining data and real-world point clouds. Current methods suffer from structural discrepancies when transferring object priors from synthetic datasets (\eg{}, ShapeNet) to real scans (\eg{}, ScanNet), particularly due to morphological variations and occlusion artifacts. To address this, \nickname{} integrates two innovative modules: (1) An object discerning module that dynamically refines object candidates, 
enabling continuous adaptation of object priors to target domains; and (2) An object completion module that reconstructs partial geometries after discovering objects. We conduct extensive experiments on both real-world and synthetic datasets, demonstrating superior 3D object segmentation performance over all baselines while achieving state-of-the-art results.
\end{abstract}

\section{Introduction}

\begin{figure*}[t]
\centering 
\centerline{\includegraphics[width=1\textwidth]{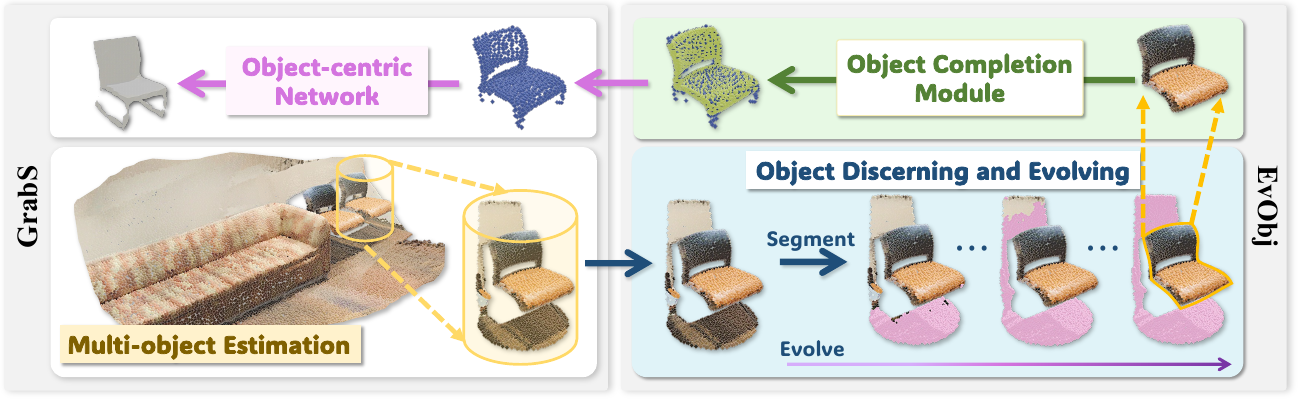}}
\vspace{-0.3cm}
    \caption{An illustration of the overall framework.}
    \label{fig:pipeline_overview}\vspace{-0.6cm}
\end{figure*}

Identifying object instances in 3D point clouds serves as a critical enabler for emerging applications, including autonomous driving, AR/VR, and robotic manipulation. Current approaches primarily follow two paradigms: (1) methods relying on high-quality 3D annotations (\eg{}, instance masks or bounding boxes) for supervised neural network training \cite{Yang2019d,Kolodiazhnyi2024}, or (2) techniques leveraging multimodal foundation models pretrained on 2D masks or image-language pairs to transfer object priors to point clouds \cite{Takmaz2023,Nguyen2024,Xu2025}. Despite achieving impressive results in closed/open-vocabulary settings, these methods require labor-intensive annotations of point clouds, images, or texts, limiting their practical utility in real-world applications.  

To eliminate scene data annotations, recent unsupervised 3D instance segmentation methods primarily follow three strategies: (1) Motion-based object discovery, which learns objectness from dynamic point cloud sequences. While achieving promising results, these methods \cite{Baur2021,Song2022,Song2024b,Zhang2023} are constrained to simple object categories (\eg{}, vehicles) due to their dependence on observable motion cues. (2) 2D-to-3D feature transfer, which lifts self-supervised image features to infer 3D objects. Though effective in open-vocabulary settings, these approaches \cite{Rozenberszki2024,Shi2024} typically yield fragmented objects due to the lack of objectness in self-supervised image features and potential misalignment between 2D and 3D features. (3) Geometric reconstruction priors, which are represented by EFEM \cite{Lei2023} and GrabS \cite{Zhang2025} that utilize a self-supervised 3D object reconstruction network to establish object priors. Among these unsupervised methods, GrabS \cite{Zhang2025} achieves very promising results in discovering complex objects in high-quality. 

The core innovation of GrabS lies in its two-stage architecture. First, an autoencoder-like network learns latent priors of objects through self-supervised reconstruction, called an object-centric network. Second, a policy network leverages the pretrained object priors via reinforcement learning (RL) to discover objects without 3D annotations. While achieving state-of-the-art performance, GrabS faces a fundamental geometric domain gap: the synthetic objects used in pretraining (\eg{}, ShapeNet \cite{Chang2015} chairs) exhibit significant structural discrepancies compared to real-world instances (\eg{}, ScanNet \cite{Dai2017} chairs). This domain shift manifests through: (1) Morphological Variations: Source domain objects often lack the topological complexity of real-world counterparts (\eg{}, simplified chair designs in ShapeNet \textit{vs} ergonomic ScanNet chairs). (2) Occlusion Artifacts: Real scans frequently contain partial geometries due to self-occlusions (\eg{}, chair legs obscuring each other), mutual occlusions (\eg{}, tables obstructing chair visibility), and sensor limitations (\eg{}, truncated object boundaries). 

In this paper, we address this core limitation of GrabS by introducing two additional modules chained together to bridge the geometric domain gap. As illustrated in the bottom-right block of Figure \ref{fig:pipeline_overview}, the first module aims to dynamically discern and \textbf{ev}olve potential \textbf{obj}ect candidates from any subset of points selected from target 3D scenes. Crucially, this continuously trainable module leverages accumulated target-domain object candidates as self-supervised pseudo-labels, enabling progressive adaptation of object priors throughout the discovery process. As shown in the top-right block of Figure \ref{fig:pipeline_overview}, the second module aims to complete partial object candidates before feeding them into the pretrained object-centric network for objectness scoring. This module specifically mitigates occlusion artifacts frequently encountered in real-world scans. Our method is named \textbf{\nickname{}} and Figure \ref{fig:opening} shows qualitative results from an indoor 3D scene. Our key contributions are:
\begin{itemize}[leftmargin=*]
\setlength{\itemsep}{1pt}
\setlength{\parsep}{2pt}
\setlength{\parskip}{1pt}  
    \item We introduce an object candidate evolution module that continuously refines target-domain candidates through self-supervised pseudo-labeling, enabling progressive adaptation of object priors during discovery. 
    \item We demonstrate significant improvements over all existing unsupervised methods across multiple benchmarks, pushing the boundaries of what's achievable without 3D scene supervision. Our code and data are available at \mbox{\url{https://github.com/vLAR-group/EvObj}}
\end{itemize}

\section{Related Works}

\phantom{xw}\textbf{3D Supervised Object Segmentation}: 
The availability of large-scale 3D datasets with high-quality human annotations (\eg{}, ScanNet \cite{Dai2017} and S3DIS \cite{Armeni2017}) has enabled significant progress in fully-supervised 3D instance segmentation. These approaches span detection-based frameworks \cite{Hou2019,Yi2019,Yang2019d,He2021,Shin2024}, clustering techniques \cite{Wang2018d,Han2020,Chen2021,Vu2022}, and Transformer architectures \cite{Schult2023,Lai2023,JiahaoLu2023,Sun2023,Kolodiazhnyi2024} to learn object masks. To reduce the annotation burden, weakly-supervised methods leverage sparse annotations such as object centers \cite{Griffiths2020} or bounding boxes \cite{Tang2022,Chibane2022,Deng2025}. While achieving notable accuracy on benchmark datasets, both paradigms fundamentally depend on costly human labels, constraining their scalability for real-world deployment.   

\textbf{Multimodal Supervised 3D Object Segmentation}: Recent advances in multimodal foundation models (\eg{}, CLIP \cite{Radford2021}, SAM \cite{Kirillov2023}, LLaVA \cite{Liu2023b}) have enabled numerous methods \cite{Ha2022,Lu2023,Takmaz2023,Liu2023,Roh2024,Yan2024,Nguyen2024,Huang2024,Yin2024,Guo2024,Boudjoghra2025,Zhao2025,Xu2025,Nguyen2025,Jung2025} to project pretrained 2D visual or vision-language features into 3D space for object discovery, often extending to open-vocabulary settings. While these approaches demonstrate impressive cross-modal transfer capabilities, they fundamentally inherit their dependency on massive human-annotated 2D datasets (image masks, captions, or aligned image-text pairs) used during foundation model pretraining. This persistent annotation requirement ultimately constrains their practicality for real-world 3D understanding applications where labeled data are scarce.

\textbf{Unsupervised 3D Object Segmentation}: To circumvent the need for scene annotations, emerging unsupervised approaches include three primary paradigms: (1) Motion-based discovery: These methods \cite{Baur2021,Song2022,Song2024b,Zhang2023} leverage motion cues in dynamic point cloud sequences to identify objects. While effective in autonomous driving contexts for moving entities like cars, their reliance on observable motion fundamentally limits applicability to static indoor environments where most objects remain stationary. (2) Transferring self-supervised 2D features: These approaches  \cite{Rozenberszki2024,Shi2024} typically project self-supervised 2D features such as DINO/2 \cite{Caron2021,Oquab2024} into 3D space followed by point clustering. Though capable of open-vocabulary discovery, they often yield fragmented segments due to the inherent lack of true objectness in self-supervised 2D features as revealed in \cite{Yang2025}. (3) Leveraging 3D reconstruction priors: Represented by recent methods like EFEM \cite{Lei2023} and GrabS \cite{Zhang2025}, this paradigm harnesses geometric reconstruction networks to establish object-centric priors. We contend that advances in large-scale 3D generative models offer particularly promising pathways for transferring rich 3D object-centric knowledge to unsupervised segmentation in complex scenes.

\begin{figure*}[t]
\centering
\includegraphics[width=1.\linewidth]{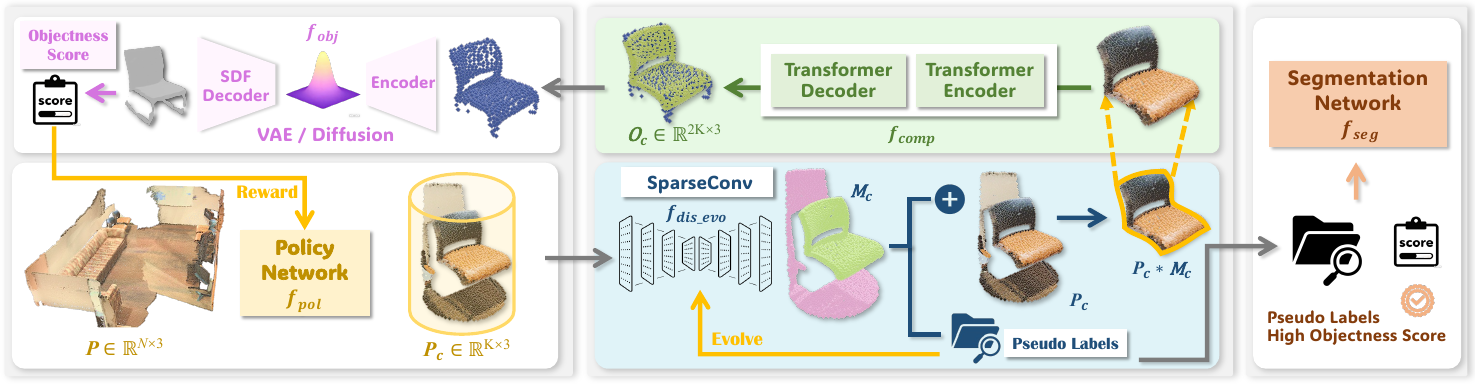}
\vskip -0.1in
\caption{Overview of our proposed pipeline. The middle blue block shows our object candidate discerning and evolving module, and the middle green block shows our object candidate completion module. The other three blocks follow the design of GrabS.}
\label{fig:method}
\vspace{-0.4cm}
\end{figure*}

\section{\nickname{}}

Our method extends the recent GrabS \cite{Zhang2025} method which successfully demonstrates object-centric priors pretrained on synthetic object datasets like ShapeNet \cite{Chang2015} can be effectively used to discover 3D objects in complex 3D scenes without needing additional human annotations to train neural networks. We first briefly review the GrabS architecture in Section \ref{sec:bg_grabs}, and then clarify our design of two modules, object candidate discernment and evolvement in Section \ref{sec:evo} and object completion in Section \ref{sec:comp}.  

\subsection{Background: GrabS}\label{sec:bg_grabs}

As illustrated in the left block of Figure \ref{fig:method}, GrabS comprises two core components:

\textbf{Object-centric Network $f_{obj}$}: It aims to learn object latent priors by training a generative reconstruction network. Particularly, given a 3D object, the object-centric network takes its downsampled point cloud as input, and then adopts either a VAE \cite{Kingma2014} or Diffusion \cite{Ho2020} framework to learn a 128-dimensional object latent distribution, followed by MLPs to decode an SDF \cite{Park2019} for the input object. 

By training on an object dataset like ShapeNet \cite{Chang2015}, the object-centric network easily learns latent shape priors for 3D objects. During testing, if a non-object 3D shape is fed into this object-centric network, the corresponding latent code is likely out of distribution, resulting in the decoded SDF being an invalid object. In this way, the trained object-centric network can serve as an indicator of objectness, which will be used in its second component.     

\textbf{Object Discovery Policy Network $f_{pol}$}:
It aims to discover similar objects on complex 3D scene point clouds without needing additional human annotations for training neural networks. Specifically, given a single scene point cloud as input, this component learns a policy network to control a dynamic container agent (\eg{}, a cylinder) to crop a point set as an object candidate or proposal. This candidate will be fed into the pretrained (frozen) object-centric network for scoring its objectness. If it has a higher score, it means the candidate is more likely to be a valid object. This objectness score serves as a reward to establish an RL pipeline to optimize the network. During training this policy network, discovered object candidates with higher objectness scores will be accumulated, serving as pseudo labels for training a separate feedforward segmentation network $f_{seg}$. More details are in GrabS \cite{Zhang2025}. 

The key issue faced by GrabS is the gap between objectness priors learned from an object dataset like ShapeNet and the complex objects in real-world scene-level point cloud scans. To address this, we significantly enhance the learning of object-centric priors by proposing two modules as detailed in Sections \ref{sec:evo}\&\ref{sec:comp}.

In this paper, we directly reuse the pretrained object-centric network from GrabS as it is, and adopt its original design of the policy network for a dynamic container agent to discover object candidates. Then, our key objective here is to bridge the gap between policy network and the pretrained object-centric network.

\subsection{Discerning and Evolving Object Candidates}\label{sec:evo}

As illustrated in Figure \ref{fig:method} (yellow block), given an input 3D scene point cloud $\boldsymbol{P}\in\mathbb{R}^{N\times 3}$, when training the policy network, the dynamic container agent is designed to crop a set of 3D points from $\boldsymbol{P}$. Such a set of 3D points is often a mixture of background points and object parts. Obviously, directly feeding such a set of points into the pretrained object-centric network $f_{obj}$ is notoriously challenging to produce a correct objectness score, leading to inaccurate rewards for the policy network, ultimately misleading object discovery. 

Fundamentally, the reasons are threefold. First, the pretrained reconstruction based object-centric network $f_{obj}$ lacks the discrimination ability to automatically ignore the noisy background points usually from floors or walls. Second, the object candidate within the point set may exhibit different morphological variations, but there is a lack of a mechanism to track and adapt to shape variations during the whole object discovery process continuously. Third, the object candidate is often incomplete due to occlusions, but the pretrained $f_{obj}$ lacks the ability to complete it. 

To this end, we develop the first module to particularly discern and evolve object candidates from any set of points selected by the policy network, and the second module to complete object candidates as detailed in Section \ref{sec:comp}.

As shown in Figure \ref{fig:method} (blue block), given a subset of points $\boldsymbol{P}_c\in \mathbb{R}^{K\times3}$ selected by the dynamic container agent from the input point cloud $\boldsymbol{P}$, where $K$ denotes the number of points within the dynamic container, we directly feed it into our object candidate discerning and evolving module, which is a per-point binary segmentation network $f_{dis\_evo}$: 
\begin{equation}\vspace{-0.2cm}
    \boldsymbol{M}_c = f_{dis\_evo}(\boldsymbol{P}_c), \quad \boldsymbol{M}_c \in \mathbb{R}^{K\times1}
\end{equation}
where $\boldsymbol{M}_c$ represents the foreground object candidate shape mask. 
For simplicity, we adopt SparseConv \cite{Graham2018} as the network architecture. Importantly, to achieve the desired object candidate discerning and evolving, the network is trained by the following two stages:

\textbf{Stage \#1 - Pretraining}: 
$f_{dis\_evo}$ is first pretrained on a synthetic object dataset like ShapeNet. By randomly adding planes to emulate walls and floors, we augment single object point clouds and create training data pairs, \ie{}, noisy object point clouds and the foreground object labels. After pretraining, $f_{dis\_evo}$ can effectively discern a foreground object from a noisy input point set, but its capability is limited to simple synthetic shapes.

\textbf{Stage \#2 - Evolving:} Given the pretrained $f_{dis\_evo}$, for any point subset $\boldsymbol{P}_c$ selected by the dynamic container agent, a reasonable foreground mask $\boldsymbol{M}_c$ will be estimated immediately, followed by feeding it into the object candidate completion module (Figure \ref{fig:method} green block), and then into the pretrained object-centric network for scoring objectness (Figure \ref{fig:method} pink block). With more and more foreground masks with higher objectness scores accumulated, denoted by $\{\boldsymbol{M}_c^1 \cdots \boldsymbol{M}_c^h \cdots \boldsymbol{M}_c^H\}$, we will reuse them as pseudo labels to finetune the network $f_{dis\_evo}$. Particularly, while training the policy network of the dynamic container agent, after every $T$ epochs, we will only use the latest accumulated foreground masks to finetune $f_{dis\_evo}$. Once used, these pseudo labels will be discarded. This simple strategy allows us to continuously adapt the object candidate discerning network to new object variations appearing in the complex scene-level point clouds, instead of statically focusing on pretrained synthetic shapes. More details of this module are in Appendix \ref{app:obj_dis_evo}.

\subsection{Completing Object Candidates}\label{sec:comp}

Following our first module designed in Section \ref{sec:evo}, given a subset of points $\boldsymbol{P}_c$, we now have its foreground object candidate mask $\boldsymbol{M}_c$, which is often incomplete due to various occlusions in real-world point clouds. As object priors in the pretrained object-centric network are learned from synthetic shapes, these incomplete object candidates can only receive rather low objectness scores, misleading the policy network to discover object candidates.  

To this end, we simply feed the foreground object candidate into a completion network $f_{comp}$ (Figure \ref{fig:method} green block), predicting a full shape, denoted by $\boldsymbol{O}_c$:
\begin{equation}\vspace{-0.2cm}
    \boldsymbol{O}_c = f_{comp}(\boldsymbol{P}_c*\boldsymbol{M}_c), \quad \boldsymbol{O}_c \in \mathbb{R}^{K\times3}
\end{equation}

For simplicity, we adopt the existing architecture AdaPoinTr \cite{Yu2023} and pretrain this completion network on ShapeNet objects from scratch. In particular, for a 3D object in ShapeNet, a partial point cloud (one or a few depth views combined) is fed into $f_{comp}$ to predict the full 3D shape. More details of this module are in Appendix \ref{app:obj_comp}.

\subsection{Training and Inference}\label{sec:train_test}

\textbf{Training}: The entire framework is trained via Algorithm \ref{alg:training}.
\vspace{-0.2cm}

\noindent\textbf{Inference}:
During testing, given a scene point cloud, we directly use the well-trained segmentation network $f_{seg}$ (Figure \ref{fig:method} orange block) to predict object masks.

\begin{figure*}[t]
\centering 
\centerline{\includegraphics[width=.95\textwidth]{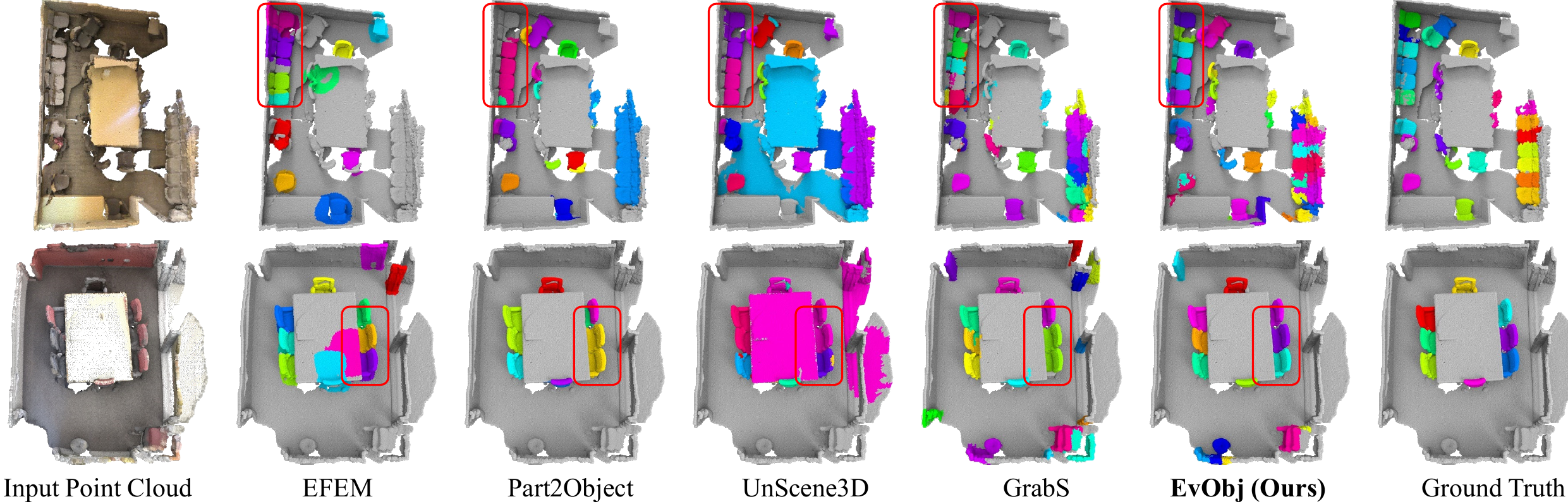}}
\vspace{-0.2cm}
    \caption{Qualitative results in the ScanNet validation set. Red boxes highlight the differences.}
    \label{fig:scannet_res}
    \vspace{-0.5cm}
\end{figure*}

\begin{algorithm}[t]
\caption{The training pipeline for the entire framework.}
\begin{algorithmic}
\label{alg:training}
\STATE{\textbf{Object-centric Models to Pretrain:} 1) the object-centric network $f_{obj}$ pretrained on ShapeNet by GrabS; 2) the object candidate discerning and evolving module $f_{dis\_evo}$ pretrained on ShapeNet; 3) the object candidate completion module $f_{comp}$ pretrained on ShapeNet. \\
}
\STATE{\textbf{Target Models to Train/Evolve:} 1) the policy network $f_{pol}$; 2) the object candidate evolving module $f_{dis\_evo}$; 3) the final feedforward segmentation network $f_{seg}$.}

\STATE{\textbf{Input:} a dataset of 3D scene point clouds without human labels: $\{\cdots, \boldsymbol{P}, \cdots\}$, where $\boldsymbol{P}$ is a single point cloud.
}
     \STATE $\bullet$ Feed $\boldsymbol{P}$ into $f_{pol}$, getting a subset of points $\boldsymbol{P}_c$;
     \STATE $\bullet$ Feed $\boldsymbol{P}_c$ into $f_{dis\_evo}$, and then into $f_{comp}$, getting foreground object candidate $\boldsymbol{O}_c$; 
     \STATE $\bullet$ Feed $\boldsymbol{O}_c$ into $f_{obj}$ to obtain an objectness score;
     \STATE $\bullet$ Use scores as rewards to optimize $f_{pol}$ by PPO;
     \STATE $\bullet$ Use object candidates with high scores as pseudo labels to finetune $f_{dis\_evo}$, after every $T$ epochs;
     \STATE $\bullet$ Ultimately, use all accumulated object candidates with high scores as pseudo labels to train the feedforward segmentation network $f_{seg}$, exactly following GrabS. 
\end{algorithmic}
\end{algorithm}

\section{Experiments}\label{sec:exp}

\textbf{Datasets}: We evaluate our method on both real-world and synthetic datasets, detailed as follows:
\textbf{ScanNet} \cite{Dai2017}: It is a challenging dataset due to various occlusions, distortions, and noisy data, including 1201, 312, and 100 scenes for training, validation, and online test, respectively.
\textbf{S3DIS} \cite{Armeni2017}: It includes six large areas of indoor scenes.
\textbf{Synthetic Multi-class Dataset}: Following GrabS \cite{Zhang2025}, we create 4000 training scenes and 1000 test scenes using multi-class objects from ShapeNet. We then replace these objects with occluded counterparts to simulate view absence in real-scanned data. Details of the datasets are in Appendix \ref{app:syn_construct}.

\textbf{Baselines}: We compare our method with the following relevant approaches:
\textbf{GrabS} \citep{Zhang2025}: It formulates unsupervised object segmentation as a two-stage pipeline, effectively identifying complex objects in 3D scenes.
\textbf{EFEM} \citep{Lei2023}: It learns object priors from ShapeNet and segments objects via EM optimization, without requiring scene annotations.
\textbf{UnScene3D} \citep{Rozenberszki2024}: It leverages pre-trained features from CSC \citep{fang2023explore} and DINO \citep{Caron2021}, performing clustering on these features to obtain pseudo labels for training an unsupervised object segmentation model.
\textbf{Part2Object} \citep{Shi2024}: It projects pixel-level pseudo masks to 3D points with the aid of DINOv2 features \cite{Oquab2024}.

\textbf{Metrics}: We evaluate the performance of class-agnostic object segmentation using standard average precision (\textbf{AP}) scores following the strategy of widely used benchmark \cite{Dai2017}. The AP scores at IoU thresholds of 25\% (\textbf{AP@25}), 50\% (\textbf{AP@50}), and averaged over all thresholds (\textbf{AP}) ranging from 50\% to 95\% at an interval of 5\% are reported.

\subsection{Evaluation on ScanNet Dataset}\label{sec:exp_scannet}
The ScanNet is widely used for evaluating the 3D indoor object segmentation task. Following GrabS \citep{Zhang2025}, we reuse its pre-trained object-centric network and train the completion network on ShapeNet chair data \citep{Chang2015}. The discerning module is optimized and evolves during training the object discovery policy network on ScanNet training set without any annotations. After training, our \nickname{} is evaluated exclusively on chairs in both validation and the online hidden test sets, treating all output masks as chairs.

\textbf{Analysis}: 
Quantitative results are reported in Tables \ref{tab:exp_scannet}\&\ref{tab:exp_scannet_test}, and our method clearly outperforms all unsupervised baselines by large margins on both validation and test sets. In the online test set, our \nickname{} even achieves almost the same segmentation performance as 3D-BoNet \cite{Yang2019d}, which is a supervised training framework, indicating \nickname{} can accurately identify target objects.
Figure \ref{fig:scannet_res} compares the visualization results of predicted masks from \nickname{} and baselines. We can see that our \nickname{} clearly detects objects that are missed by baselines. The reason is that we can narrow the domain gap between synthetic dataset and ScanNet. Detailed analyses are provided in Section \ref{exp:analysis}, and more results are shown in Appendix \ref{app:scannet}.
\begin{table}[th]\tabcolsep= 0.15cm 
\centering
 \setlength{\abovecaptionskip}{ 2 pt}
\caption{Quantitative results of our method and baselines on the ScanNet validation set \cite{Dai2017}.}
\label{tab:exp_scannet}
\resizebox{1.0\linewidth}{!}
{
\begin{tabular}{crccc}
\toprule[1.0pt]
& & AP(\%) & AP@50(\%)& AP@25(\%) \\
\toprule[1.0pt]
\multirow{1}{*}{\makecell[c]{Supervised}} 
& Mask3D \cite{Schult2023}& 82.9 & 94.4 & 97.0 \\
\toprule[1.0pt]
\multirow{6}{*}{\makecell[l]{Unsupervised}}
& UnScene3D \cite{Rozenberszki2024} & 37.2 & 62.4 & 79.2\\
& Part2Object \cite{Shi2024}& 34.4 & 56.8 & 73.9\\
& EFEM \cite{Lei2023} & 24.6 & 50.8 & 61.3\\
& GrabS-VAE \cite{Zhang2025} & 46.7 &71.5 &82.9\\
& GrabS-Diffusion \cite{Zhang2025} &47.1 & 70.6 & 81.1\\
&\textbf{\nickname{} (Ours-VAE)} &\textbf{55.0}  &\textbf{76.9}  &88.2 \\
&\textbf{\nickname{} (Ours-Diffusion)} &54.7  &76.0  &\textbf{88.6} \\
\bottomrule[1.0pt]
\end{tabular}
}
\vspace{-0.6cm}
\end{table}

\begin{table}[ht]
\tabcolsep= 0.1cm 
\centering
 \setlength{\abovecaptionskip}{ 2 pt}
\caption{Quantitative results of our method and baselines on the hidden test set of ScanNet.}
\label{tab:exp_scannet_test}
\resizebox{1.0\linewidth}{!}
{
\begin{tabular}{lrccc}
\toprule[1.0pt]
& & AP(\%) & AP@50(\%)& AP@25(\%) \\
\toprule[1.0pt]
\multirow{3}{*}{\makecell[l]{Supervised} }
& 3D-BoNet \citep{Yang2019d}& 34.5 & 48.4 & 64.3 \\
& SoftGroup \citep{Vu2022} & 69.4 & 86.2 & 91.3\\
& Mask3D \citep{Schult2023}& 73.7 & 88.5 & 93.8 \\
\toprule[1.0pt]
\multirow{3}{*}{\makecell[l]{Unsupervised}} 
& EFEM \citep{Lei2023} &20.2  &39.0 &48.3 \\
& GrabS-VAE \cite{Zhang2025} &29.0 &45.1 &57.7 \\
& GrabS-Diffusion \cite{Zhang2025} &28.5  &43.1 &58.1 \\
&\textbf{\nickname{} (Ours-VAE)} &\textbf{34.0}  &47.6  &\textbf{59.7} \\
&\textbf{\nickname{} (Ours-Diffusion)} &33.9  &\textbf{47.8}  &58.1 \\

\bottomrule[1.0pt]
\end{tabular}
}
\vspace{-0.6cm}
\end{table}

\begin{figure*}[ht]
\centering 
\centerline{\includegraphics[width=1\textwidth]{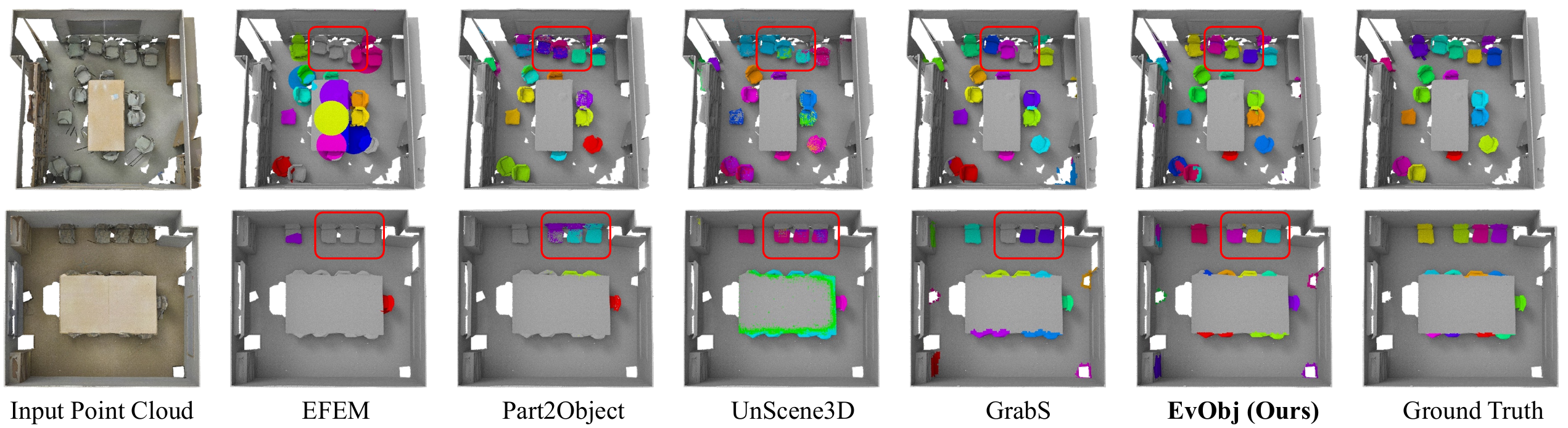}}
\vspace{-0.2cm}
    \caption{Qualitative results on the S3DIS dataset. Red boxes highlight the differences.}
    \label{fig:S3DIS_res}
    \vspace{-0.5cm}
\end{figure*}

\subsection{Generalization to S3DIS Dataset}\label{sec:exp_s3dis}

To verify the generalization capability of our \nickname{}, we follow Part2Object \citep{Shi2024} and GrabS \citep{Zhang2025} to conduct experiments under a cross-dataset validation setting. Specifically, we reuse the well-trained segmentation models from the ScanNet and evaluate them on the S3DIS dataset.

\textbf{Analysis}: 
Tables \ref{tab:exp_s3dis_area5}\&\ref{tab:exp_s3dis_6fold} demonstrate the performance of baselines and \nickname{}. Our method achieves higher AP scores across all areas, significantly outperforming UnScene3D \citep{Rozenberszki2024} and Part2Object \citep{Shi2024}. The qualitative results in Figure \ref{fig:S3DIS_res} illustrate that \nickname{} can identify target objects in crowded regions, owing to the dynamic evolving of the object candidates discerning module. More results and visualizations on all six areas are provided in Appendix \ref{app:s3dis}.

\begin{table}[th]\tabcolsep= 0.15cm 
\centering
 \setlength{\abovecaptionskip}{ 2 pt}
\caption{Quantitative results of our method and baselines on the S3DIS-Area5 \cite{Dai2017}.}
\label{tab:exp_s3dis_area5}
\resizebox{1.0\linewidth}{!}
{
\begin{tabular}{crccc}
\toprule[1.0pt]
& & AP(\%) & AP@50(\%)& AP@25(\%) \\
\toprule[1.0pt]
\multirow{6}{*}{\makecell[l]{Unsupervised}}
& UnScene3D \cite{Rozenberszki2024} & 42.6 & 63.4 &80.3 \\
& Part2Object \cite{Shi2024}& 30.0 & 50.5 & 76.4\\
& EFEM \cite{Lei2023} & 14.9 & 35.7 & 45.3 \\
& GrabS-VAE \cite{Zhang2025}&46.4 &66.2 &73.8\\
& GrabS-Diffusion \cite{Zhang2025} & 44.2 & 58.0 & 62.6\\
&\textbf{\nickname{} (Ours-VAE)} &55.1  &77.1  &86.2 \\
&\textbf{\nickname{} (Ours-Diffusion)} &\textbf{60.6}  &\textbf{82.8}  &\textbf{92.1} \\
\bottomrule[1.0pt]
\end{tabular}
}
\vspace{-0.6cm}
\end{table}

\begin{table}[th]\tabcolsep= 0.15cm 
\centering
 \setlength{\abovecaptionskip}{ 2 pt}
\caption{Quantitative results of our method and baselines on the S3DIS 6-fold \cite{Dai2017}.}
\label{tab:exp_s3dis_6fold}
\resizebox{1.0\linewidth}{!}
{
\begin{tabular}{crccc}
\toprule[1.0pt]
& & AP(\%) & AP@50(\%)& AP@25(\%) \\
\toprule[1.0pt]
\multirow{6}{*}{\makecell[l]{Unsupervised}}
& UnScene3D \cite{Rozenberszki2024}& 30.3 & 51.9 &70.4\\
& Part2Object \cite{Shi2024} & 25.3 & 48.4 & 67.0 \\
& EFEM \cite{Lei2023} & 16.2 & 37.8 & 45.9 \\
& GrabS-VAE \cite{Zhang2025} &41.8 &61.7 & 67.0  \\
& GrabS-Diffusion \cite{Zhang2025} & 39.2 & 57.2 & 62.6  \\
&\textbf{\nickname{} (Ours-VAE)} &46.3 &67.5 &77.4 \\
&\textbf{\nickname{} (Ours-Diffusion)} &\textbf{47.4} &\textbf{69.3} &\textbf{78.6} \\
\bottomrule[1.0pt]
\end{tabular}
}
\vspace{-0.6cm}
\end{table}

\subsection{Evaluation on Multi-class Dataset}\label{sec:exp_multi}
\begin{figure*}[t]
\centering 
\centerline{\includegraphics[width=1\textwidth]{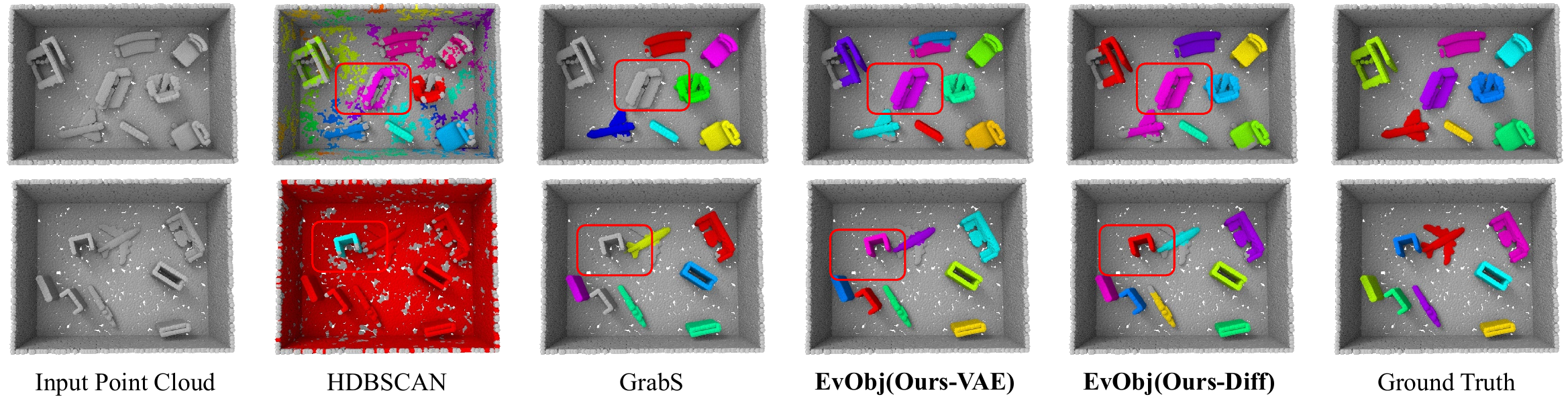}}
\vspace{-0.2cm}
    \caption{Qualitative results on the test set of our synthetic multi-class dataset.}
    \label{fig:synthetic_res}
    \vspace{-0.45cm}
\end{figure*}

Real-world scenes contain objects of various categories and often suffer from self-/mutual-occlusions. However, real-scanned datasets like ScanNet share only a few categories with object datasets \eg{}, ShapeNet, making the learned priors unavailable for searching all objects in them.

To validate the ability of discovering multi-class objects, we follow GrabS \citep{Zhang2025} to simulate a challenging multi-class dataset, which consists of multiple categories of ShapeNet objects with occlusions.
Specifically, we use objects from \textit{chair, sofa, telephone, airplane, rifle, and cabinet} with only a few views combined to create 5000 indoor scenes, and split them at a 4:1 ratio for training and test. All objects in our training and test sets are retrieved from the validation and test sets of ShapeNet, respectively, with layout arrangements identical to those in GrabS. More details of our synthetic dataset are provided in Appendix \ref{app:syn_construct}.

The object-centric networks of both GrabS \citep{Zhang2025} and our \nickname{} are trained on ShapeNet objects from the selected six categories. The 3D object segmentation networks are trained and evaluated on our newly created synthetic indoor scenes without any scene-level annotations. 

\textbf{Analysis}: As shown in Table \ref{tab:exp_syn} and Figure \ref{fig:synthetic_res}, our approach effectively segments partial objects belonging to multiple categories, clearly outperforming all baselines.  Our improvement is primarily due to the object candidate completion module, which allows the discovered partial objects to be completed before receiving valid objectness scores, confirmed by our significantly higher AP@25(\%) score.  More results are provided in Appendix \ref{app:syn}.

\begin{table}[th]\tabcolsep= 0.15cm 
\centering
 \setlength{\abovecaptionskip}{ 2 pt}
\caption{Quantitative results of our method and baselines on our synthetic multi-class dataset.}
\label{tab:exp_syn}
\resizebox{1.0\linewidth}{!}
{
\begin{tabular}{lrccc}
\toprule[1.0pt]
& & AP(\%) & AP@50(\%)& AP@25(\%) \\
\toprule[1.0pt]
\multirow{1}{*}{\makecell[c]{Supervised}} 
& Mask3D \cite{Schult2023} &78.6 &90.1 &94.9 \\
\toprule[1.0pt]
\multirow{2}{*}{\makecell[l]{Unsupervised \\ \& Real2Syn }}
& UnScene3D \cite{Rozenberszki2024} & 58.4 & 76.4 & 86.5 \\
& Part2Object \cite{Shi2024} & 58.6 & 77.6 & 87.2 \\
\toprule[1.0pt]
\multirow{6}{*}{\makecell[l]{Unsupervised}}
& HDBSCAN \cite{mcinnes2017} &14.2  & 31.7 & 44.6 \\
& GrabS-VAE \cite{Zhang2025} & 59.2 & 79.2 & 82.9 \\
& GrabS-Diffusion \cite{Zhang2025}& 59.5 &78.5 &81.8\\
&\textbf{\nickname{} (Ours-VAE)} &\textbf{62.1} &\textbf{85.6} &\textbf{90.3} \\
&\textbf{\nickname{} (Ours-Diffusion)} &61.0  &84.0  &90.1 \\
\bottomrule[1.0pt]
\end{tabular}
}
\vspace{-0.6cm}
\end{table}

\subsection{Ablation Study}\label{exp:abl}
To evaluate the effectiveness of each component and hyperparameter choices in our method, we conduct the following ablation experiments on the ScanNet validation set.

\textbf{(1) Removing Object Candidate Discerning Module}: Instead of introducing an object candidate discerning network ahead of the object-centric network, we simply filter point-wise reconstruction error to generate object candidate masks following GrabS.

\textbf{(2) Removing Evolving Optimization}: The object discerning module is fixed after pretraining on synthetic data. Stopping evolving aims to validate its necessity.

\textbf{(3) Removing Pretraining $f_{dis\_evo}$}: We train the discerning module $f_{dis\_evo}$ from scratch using accumulated foreground masks, instead of initializing it via pretraining.

\textbf{(4) Removing Object Candidate Completion}: We remove the object completion module, which recovers full object candidate shapes to facilitate object scoring.

\textbf{(5) $\sim$ (7) Different choices of evolving interval $T$}: $T$ controls the evolving frequency of object candidate discerning module $f_{dis\_evo}$. Lower $T$ means a higher evolving frequency, which may degrade performance due to the accumulated object candidates containing noise rather than being perfect. Higher $T$ indicates that the evolving is not enough, hurting the domain adaptation of $f_{dis\_evo}$.
We set $T$ as 100 in the main experiments.

\textbf{(8) $\sim$ (10) Different choices of completion model}: Our \nickname{} is agnostic to the completion model $f_{comp}$, and we evaluate \nickname{} with three recent choices. 

\textbf{Analysis}: From Table \ref{tab:ablative}, we can see that:
1) The proposed $f_{dis\_evo}$ is critical to the pipeline, making the scoring of the object-centric network more accurate, enabling more effective guidance. Additionally, synthetic pretraining combined with evolving bootstraps the effectiveness of $f_{dis\_evo}$.
2) Removing the object completion module incurs a clear drop in performance, showing its necessity.
3) Our \nickname{} is robust to hyperparameters in evolving and completion.

\begin{table}\tabcolsep= 0.12cm 
\centering
 \setlength{\abovecaptionskip}{ 2 pt}
\caption{The AP scores of all ablated networks on the validation set of ScanNet based on our full \nickname{}.}
\label{tab:ablative}
\resizebox{1.0\linewidth}{!}{
\begin{tabular}{lccc}
\toprule[1.0pt]
 &AP(\%) &AP@50(\%) &AP@25(\%)\\
\toprule[1.0pt]
\textbf{Ablations on Modules}\\
(1) Removing Object Discerning &45.0  &72.0  &86.2 \\
(2) Removing Evolving $f_{dis\_evo}$ &52.2  &75.6  &87.7 \\
(3) Removing Pretraining $f_{dis\_evo}$ &37.4  &62.5  &79.8 \\
(4) Removing Object Completion &33.8  &44.3  &49.2 \\
\toprule[1.0pt]
\textbf{Ablations on Evolving}\\
(5) $T=50$  &52.6  &73.4  &85.4 \\
\textbf{(6) $T$ = 100}  &\textbf{55.0}  &\textbf{76.9}  &\textbf{88.2} \\
(7) $T=200$  &53.4  &75.3  &86.5 \\
\toprule[1.0pt]
\textbf{Ablations on Completion}\\
\textbf{(8) \nickname{}$_{AdaPointTr\cite{Yu2023}}$}  &\textbf{55.0} &\textbf{76.9}  &\textbf{88.2} \\
(9) \nickname{}$_{PoinTr\cite{yu2021pointr}}$ &53.1  &75.2  &87.1 \\
(10) \nickname{}$_{SnowflakeNet \cite{xiang2021snowflakenet}}$ &54.1 &76.2 &87.9 \\
\textbf{The full \nickname{}} &\textbf{55.0} &\textbf{76.9} &\textbf{88.2} \\
\toprule[1.0pt]
\end{tabular}
}\vspace{-0.6cm}
\end{table}

\subsection{Impact of Discerning and Evolving}\label{exp:analysis}

\begin{figure*}[t]
\centering 
\centerline{\includegraphics[width=1\textwidth]{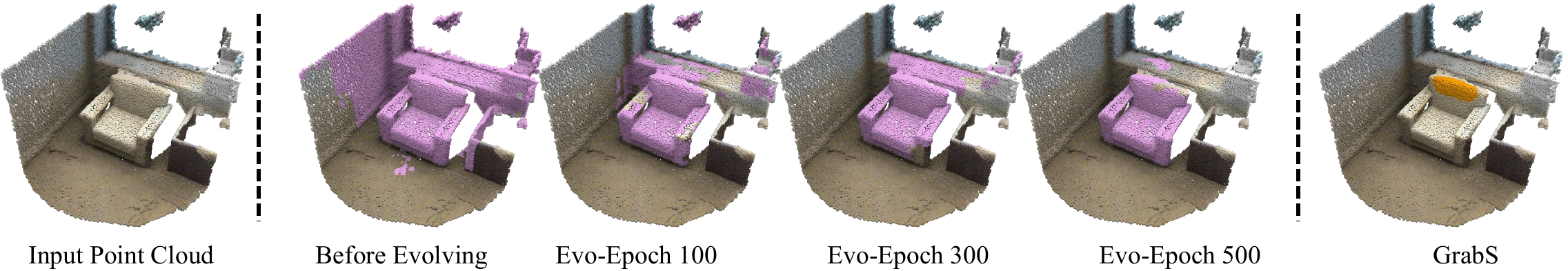}}
\vspace{-0.2cm}
    \caption{Object candidates from our discerning module during the evolving process and GrabS. After evolving, the chair candidate is gradually discovered, while the baseline GrabS fails to do so.}
    \label{fig:evo}
    \vspace{-0.4cm}
\end{figure*}

The core insight of \nickname{} is the object candidate discerning module $f_{dis\_evo}$ together with an evolving optimization strategy.
By introducing it ahead of the object-centric network, complex objects in 3D scenes $\boldsymbol{P}_{c}$ can be discerned, encouraging the agent to discover objects more accurately and comprehensively. To further evaluate the effectiveness of $f_{dis\_evo}$, we analyze the quality of progressively discovered object candidates via the RL agent, by comparing against ground truth masks.

\textbf{(1) The Sufficiency of Qualified Object Candidates}: 
A large number of object candidates are discovered by the agent during agent training. Although these candidates are assigned high scores by the object-centric network, they are not guaranteed to be true objects, as the object-centric network is not accurate. Therefore, we first filter the discovered object candidates using ground truth masks with an IoU threshold of 60\%, and candidates that meet this criterion are regarded as qualified. Second, we count the ratio of qualified object candidates among all ground truth objects, and use this ratio as a metric to measure the sufficiency of object candidates. This ratio is updated during evolving, and higher ratios indicate the agent identifies more objects.

\begin{table}[h]
\centering
\vspace{-0.1cm}
 \setlength{\abovecaptionskip}{2 pt}
\caption{The sufficiency (\%) of qualified object candidates discovered over epochs on the ScanNet training set.}
\label{tab:suff}
\resizebox{0.95\linewidth}{!}{
\begin{tabular}{crcccccc}
\toprule[1.0pt]
&Epoch &100 &200 &300 &400 &500\\
\toprule[1.0pt]
\multirow{8}{*} 
&GrabS \cite{Zhang2025} &47.9 &51.3 &52.6 &53.5 &53.7\\
&\nickname{} without $f_{dis\_evo}$ &52.5 &54.8 &55.3 &55.5 &56.2\\
&\nickname{} with $f_{dis\_evo}$ &55.1 &59.2 &60.9 &61.2 &61.3\\
\bottomrule[1.0pt]
\end{tabular}} \vspace{-0.3cm}
\end{table}

\textbf{Analysis}: The sufficiency is reported in Table \ref{tab:suff}. We can see that: 
1) With the training of agents, the discerning module $f_{dis\_evo}$ gradually adapts into the scene domain, therefore helping our \nickname{} to rapidly and correctly find more object candidates. 
2) Finally, the discerning module achieves the highest sufficiency at the ratio of 61.3\%, demonstrating that it finds objects more thoroughly, making our \nickname{} consistently outperforms GrabS \cite{Zhang2025} in all training epochs. 3) Without $f_{dis\_evo}$, \nickname{} finds fewer objects.
Figure \ref{fig:suff} shows comparisons of the extracted object candidates in complex backgrounds, which illustrates that $f_{dis\_evo}$ can discern more object variations than GrabS. More results are provided in Appendix \ref{app:analy}.

\textbf{(2) The Accuracy of Evolved Discerning Module}:
Evolving is capable of adapting the discerning module $f_{dis\_evo}$ from synthetic data to the real-scanned scene domain, leading the agent to better and better localize objects. We also measure the effects of evolving by recording the accuracy of discovered object candidates in different training epochs. 
In particular, in training the RL agent, we feed all points inside the agent container into $f_{dis\_evo}$ to predict object candidate masks and compute the accuracy by comparing against ground truth masks. Higher accuracy means the object discerning module $f_{dis\_evo}$ is more capable of extracting object points from the scene subset $\boldsymbol{P}_c$.

\textbf{Analysis}: Results in Table \ref{tab:evo} show that: 
1) With evolving, the ability of $f_{dis\_evo}$ to extract object points continuously improves, increasing from 83.2\% to 88.4\%, achieving the highest accuracy.
2) Without evolving, the discerning accuracy degrades, even slightly lower than that of GrabS. We attribute this to the better generalization ability of the generative model in GrabS \cite{Zhang2025}. Overall, the best performance is achieved by combining evolving with the object candidate discerning module. 
Qualitative results of the evolving process in Figure \ref{fig:evo} show that the evolving strategy helps discern module $f_{dis\_evo}$ to extract more and more accurate objects, significantly surpassing GrabS.

\begin{table}[h]
\centering
\vspace{-0.3cm}
 \setlength{\abovecaptionskip}{ 2 pt}
\caption{The accuracy (\%) of the object discerning module in different epochs on the ScanNet training set.}
\label{tab:evo}
\resizebox{0.95\linewidth}{!}{
\begin{tabular}{crcccccc}
\toprule[1.0pt]
&Epoch &100 &200 &300 &400 &500\\
\toprule[1.0pt]
\multirow{8}{*} 
&VAE in GrabS \cite{Zhang2025} &85.3 &85.3 &85.3 &85.3 &85.3\\
&$f_{dis\_evo}$ without evolving &84.8 &84.8 &84.8 &84.8 &84.8\\
&$f_{dis\_evo}$ with evolving &83.2 &85.4 &88.1 &87.9 &88.4\\
\bottomrule[1.0pt]
\end{tabular}}
\end{table}

\begin{figure}[h]
\centering 
 \setlength{\abovecaptionskip}{ 4 pt}
\centerline{\includegraphics[width=0.9\linewidth]{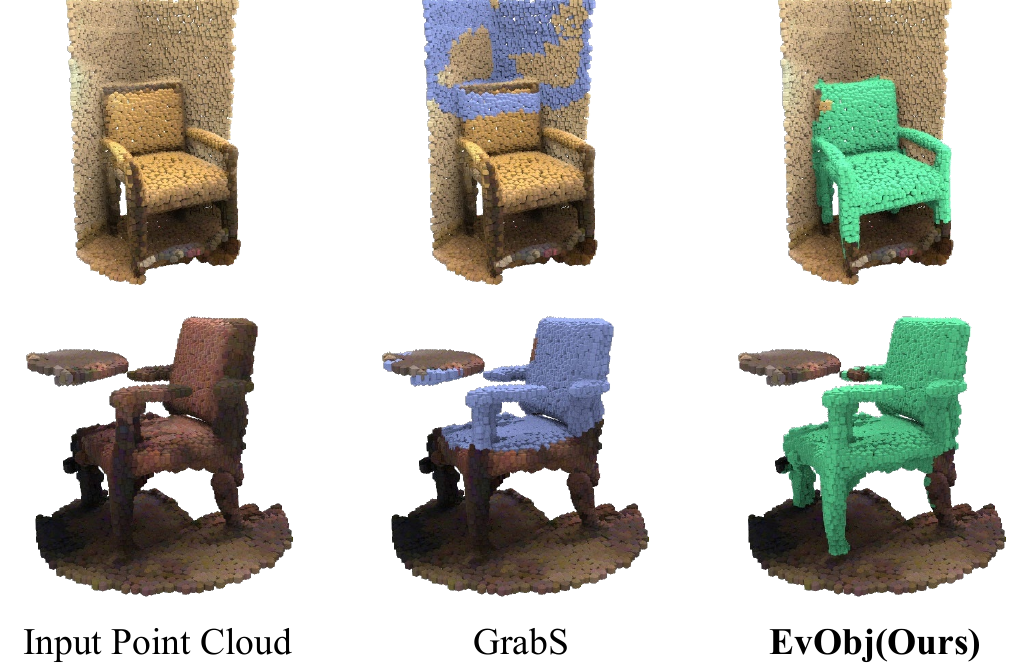}}
    \caption{Qualitative comparisons of extracted objects from subsets of the ScanNet training set.}
    \label{fig:suff}
    \vspace{-0.5cm}
\end{figure}

\section{Conclusion}

In this paper, we present \nickname{}, a method that addresses the synthetic-to-real domain gap in unsupervised 3D instance segmentation by introducing two key modules: (1) an object candidate discerning and evolving module enabling continuous adaptation to target-domain shape morphology through self-supervised pseudo-label refinement, and (2) an object candidate completion module reconstructing partial geometries before objectness scoring. Extensive experiments on multiple benchmarks demonstrate the state-of-the-art performance of our method. Notably, detailed analyses of the object discerning module reveal that it substantially improves both the sufficiency and accuracy of object candidates, contributing to the overall effectiveness of our method. In future work, we aim to explore adaptation strategies that leverage object-centric priors from large-scale object generation models and multimodal foundation models, pushing the limit of annotation-free 3D perception. 

\textbf{Acknowledgments:} This work was supported in part by National Natural Science Foundation of China under Grant 62271431, in part by Research Grants Council of Hong Kong under Grants 15219125 \& 15225522.

\clearpage
{\small
\bibliographystyle{ieeenat_fullname}
\bibliography{reference}}

\clearpage

\maketitlesupplementary

\subsection{More Details of Discerning and Evolving Object Candidates}\label{app:obj_dis_evo}
\begin{figure}[t]
\centering 
\centerline{\includegraphics[width=0.5\textwidth]{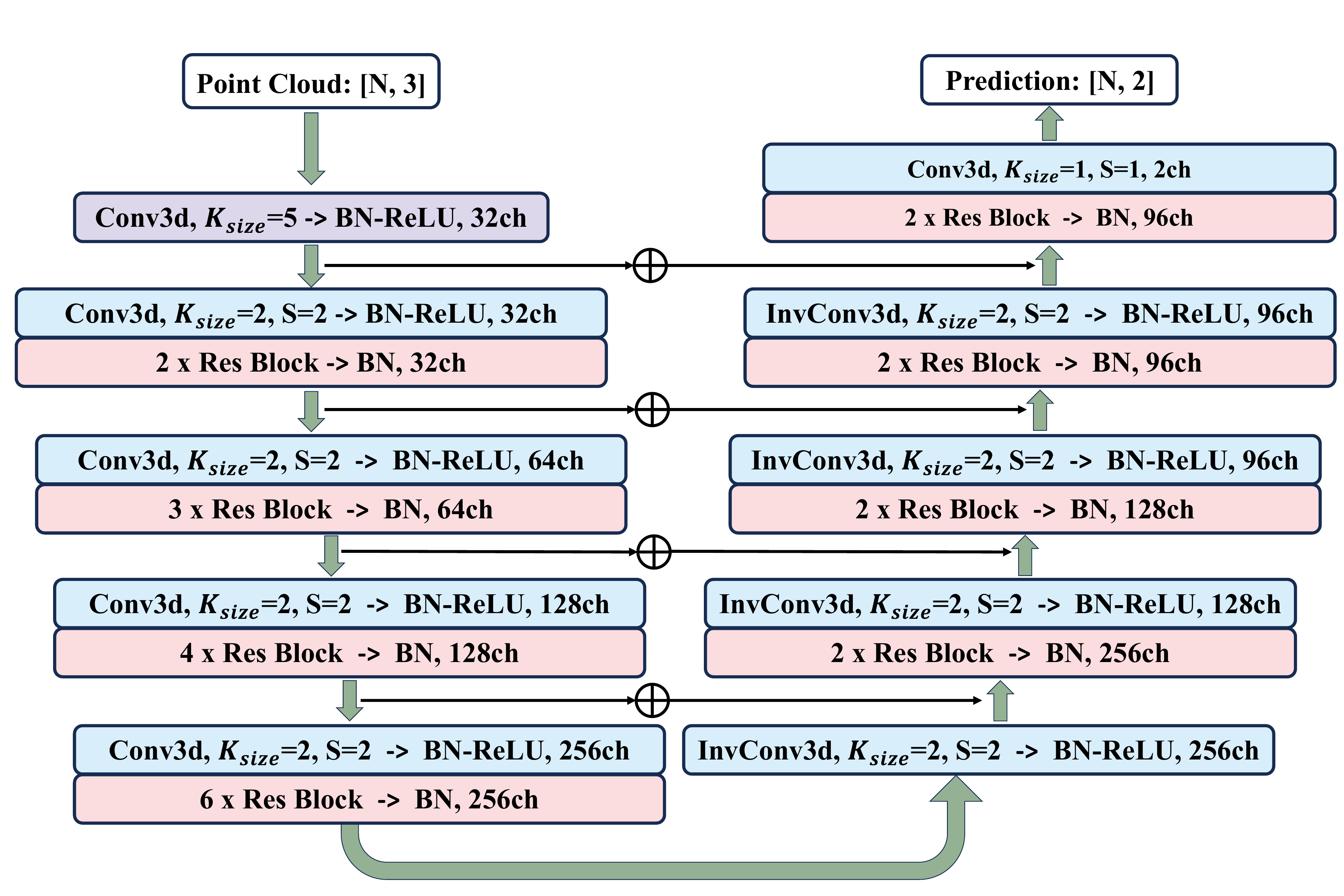}}
\vspace{-0.2cm}
    \caption{Network Structure of Discerning Module}
    \label{fig:seg_structure}
    \vspace{-0.2cm}
\end{figure}

\textbf{Network Architecture}: As shown in Figure \ref{fig:seg_structure}, the discerning module adopts SparseUNet \cite{Graham2018}. The encoder performs spatial resolution reduction through 4 stages, and decoder enhances features with upsampling. Two parallel convolution layers are concatenated at the end of UNet to produce the voxel-wise scores for foreground and background. $K_{\text{size}}$, $S$, and $ch$ in Figure \ref{fig:seg_structure} denote the kernel size, stride, and feature channels for sparse convolutions, respectively. Symbol $\oplus$ represents the concatenation of voxel features.

\textbf{Pretraining}: The discerning module is pretrained on objects from ShapeNet dataset. To simulate realistic occlusions in real-world scenarios, we render depth images for each 3D object from 12 views and randomly select and fuse 3 $\sim$ 6 views to obtain a partial point cloud. Subsequently, we generate horizontal and vertical planes in a unit cube to simulate the floor and wall, which are treated as backgrounds. Additionally, the augmented object point clouds are randomly scaled from 0.8 to 1.5. 

The pretraining of the discerning module is a point-wise binary segmentation task for distinguishing object and background points. Therefore, we adopt the cross-entropy loss function and Adam optimizer with a learning rate of 1e-3 with a weight decay of 1e-4. The training takes a batch size of 32 with a voxel size of 0.05 and lasts for 20 epochs.

\textbf{Evolving Phase}: During evolving, we gather object candidates scored by the object-centric network during reinforcement learning. Specifically, as the object-centric network is a reconstruction model, we feed points of each candidate into the network. The candidate is regarded as a valid object only when the Chamfer Distance between the input points and recovered shape is below 0.1.

The evolving is conducted every 100 epochs. We train the discerning module with the accumulated object candidates, and adopt the Adam optimizer with a learning rate of 1e-3 and a weight decay of 1e-4. Training epoch is set as 40 with a batch size of 32. Additionally, data augmentation is applied, including rotation around the z-axis from -180° to 180° and scaling within the range of 0.9 to 1.1. The accumulated object candidates will be discarded once used.

\subsection{More Details of Completing Object Candidates}\label{app:obj_comp}
Object candidates are completed through a pretrained completion model, which adopts AdaPointTr \cite{Yu2023} in our main experiments. 
However, the vanilla AdaPoinTr is trained with 3D objects in canonical pose, which is hard to apply to real-scanned data. Thus, we keep the model architecture and train it from scratch using our training data.

\textbf{Data Preparation}: We randomly combine depth scans generated in Section \ref{app:obj_dis_evo} from 2$\sim$4 views, and downsample to 1024 points as input data. The ground truth full 3D shape is constructed by randomly sampling 1024 points from the object mesh surface.  

\textbf{Training Process}: The training is kept the same as AdaPoinTr. Chamfer Distance is taken as the loss function. We train the completion model by AdamW with a learning rate of 1e-4 and weight decay of 5e-4 for 300 epochs.

\subsection{Training and Evaluation on ScanNet}\label{app:scannet}

\begin{figure*}[th]
\centering 
\centerline{\includegraphics[width=1\textwidth]{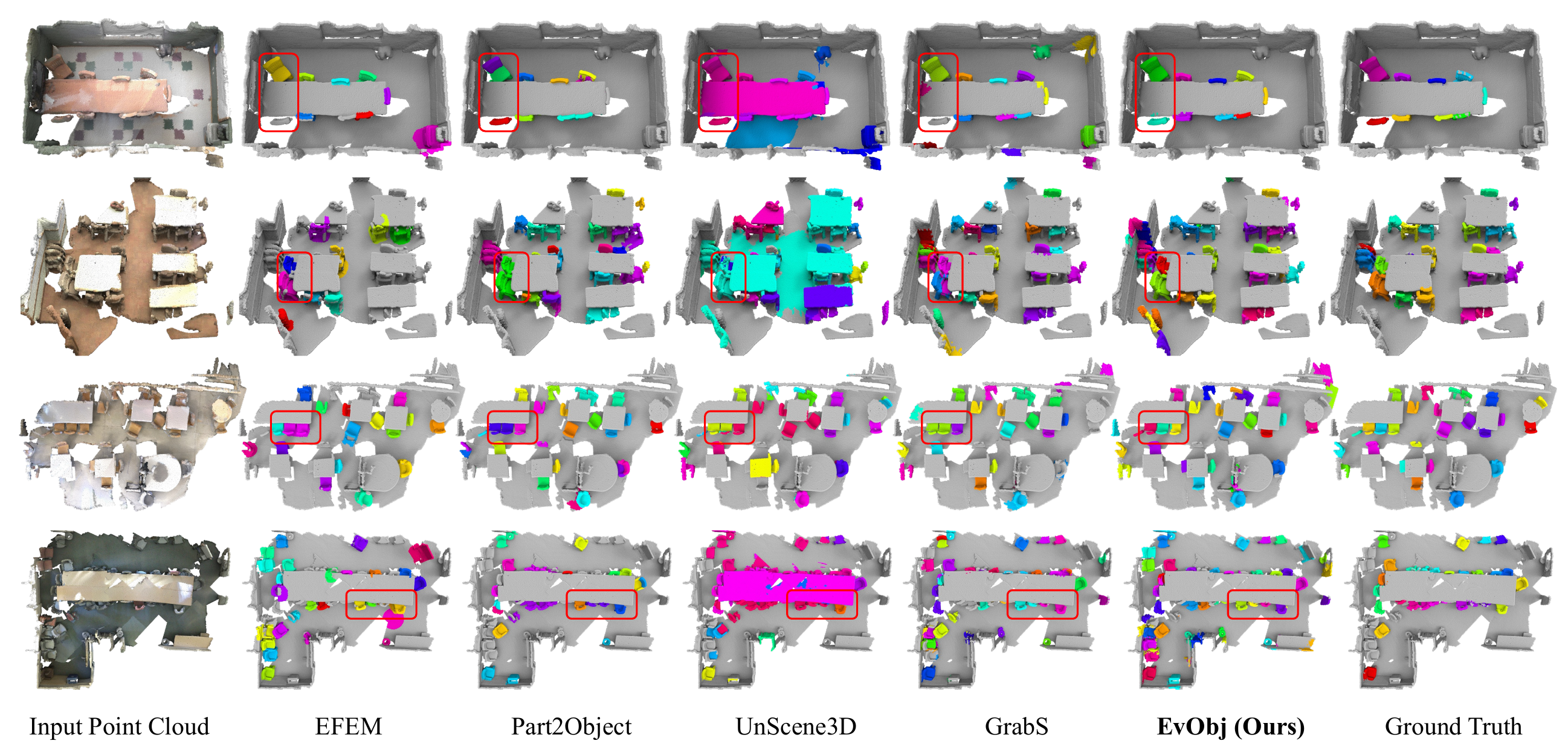}}
\vspace{-0.2cm}
    \caption{More qualitative results on ScanNet validation set. Red boxes highlight the differences.}
    \label{fig:scannet_app}
    \vspace{-0.2cm}
\end{figure*}

Following GrabS \cite{Zhang2025}, we adopt a Mask3D network structure with the SparseUNet as our backbone to extract voxel features. The segmentation head is a Transformer decoder, and the policy network is a cross-attention with two separate MLPs for predicting actions and state value. 
The object discovery networks are trained with both segmentation and PPO loss functions, which are exactly keep same as GrabS. 
Training on the ScanNet training set lasts for 600 epochs with a batch size of 8, and the AdamW optimizer is configured with a fixed learning rate of 1e-4. 
With the successful usage of superpoints in 3D unsupervised learning \cite{zhang2023growsp,zhang2025logosp,zhang2026growsp++}, we have also incorporated them into our framework. For evaluation, all masks predicted by the segmentation head are treated as chair candidates, and the Average Precision (AP) for chairs is computed on the validation and online hidden test sets of ScanNet in Table \ref{tab:exp_scannet} \&\ Table \ref{tab:exp_scannet_test}. Figure \ref{fig:scannet_app} shows the differences between our \nickname{} and the baseline on ScanNet validation set. Experimental results indicate that our method is more effective at addressing cases where multiple chairs are closely placed.

\subsection{Training and Evaluation on S3DIS}\label{app:s3dis}
\begin{figure*}[t]
\centering 
\centerline{\includegraphics[width=1\textwidth]{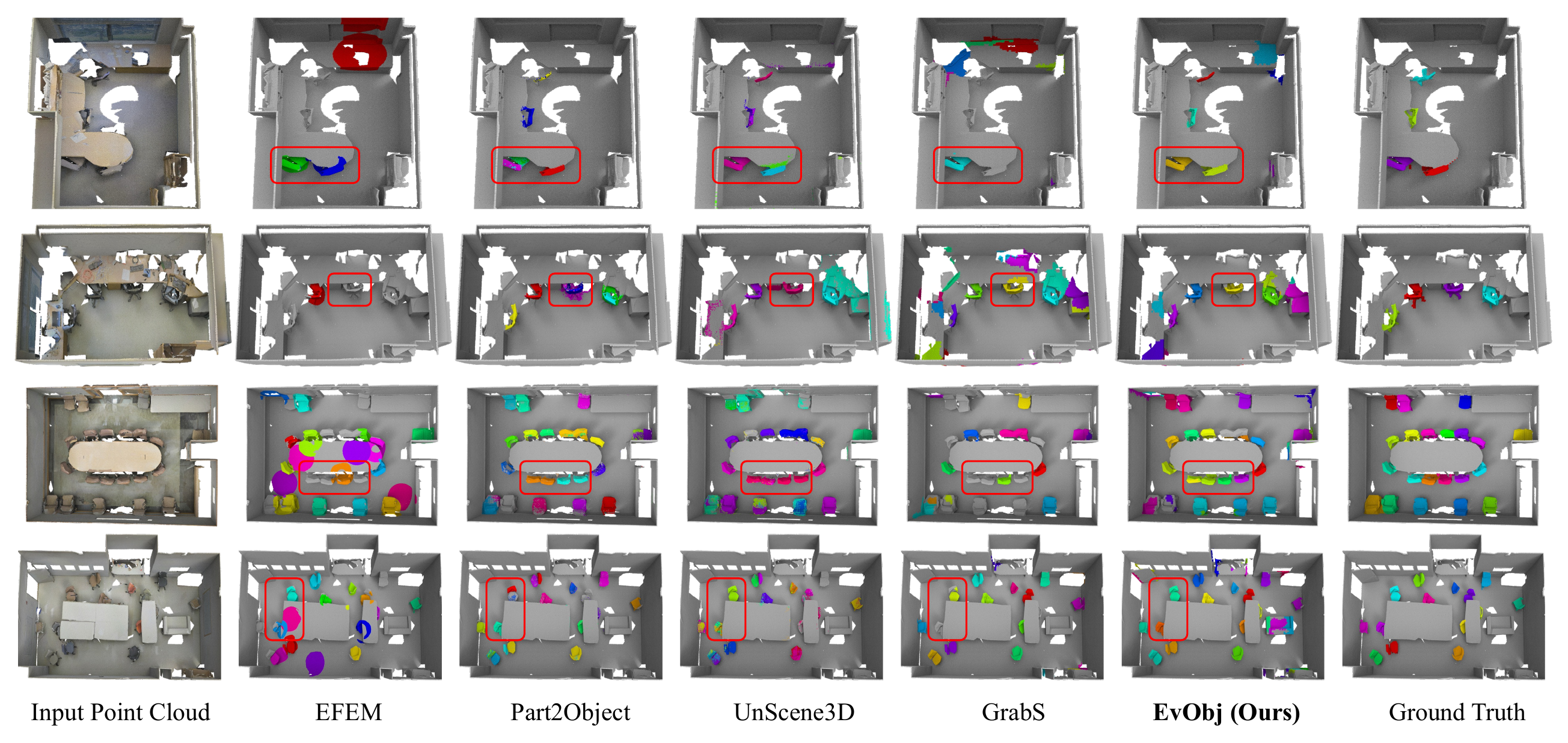}}
\vspace{-0.2cm}
    \caption{More qualitative results on S3DIS validation set. Red boxes highlight the differences.}
    \label{fig:s3dis_app}
    \vspace{-0.2cm}
\end{figure*}

The cross-dataset evaluation is conducted on the S3DIS dataset using the well-trained models from the ScanNet training set, which is the same as Part2Object and GrabS.  \cref{tab:app_s3dis_area1,tab:app_s3dis_area2,tab:app_s3dis_area3,tab:app_s3dis_area4,tab:app_s3dis_area5,tab:app_s3dis_area6} present the segmentation results on six areas in S3DIS, and more qualitative results are shown in Figure \ref{fig:s3dis_app}, where our \nickname{} significantly outperforms existing baselines.
\begin{table}[th]\tabcolsep= 0.17cm 
\centering
 \setlength{\abovecaptionskip}{ 2 pt}
\caption{Quantitative results of our method and baselines on the S3DIS-Area1 \cite{Dai2017}.}
\label{tab:app_s3dis_area1}
\resizebox{0.95\linewidth}{!}
{
\begin{tabular}{crccc}
\toprule[1.0pt]
& & AP(\%) & AP@50(\%)& AP@25(\%) \\
\toprule[1.0pt]
\multirow{6}{*}{\makecell[l]{Unsupervised}}
& UnScene3D \cite{Rozenberszki2024} &33.6  &63.8  &85.4  \\
& Part2Object \cite{Shi2024} &27.1  &55.4  &77.5  \\
& EFEM \cite{Lei2023} &19.1 &48.6 &54.7 \\
& GrabS-VAE \cite{Zhang2025} &45.5  &68.6  &73.2  \\
& GrabS-Diffusion \cite{Zhang2025} &47.8  &70.9  &75.7  \\
&\textbf{\nickname{} (Ours-VAE)} &50.7  &77.6  &86.7 \\
&\textbf{\nickname{} (Ours-Diffusion)} &\textbf{52.8}  &\textbf{82.2}  &\textbf{90.6} \\
\bottomrule[1.0pt]
\end{tabular}
}
\vspace{-0.2cm}
\end{table}

\begin{table}[th]\tabcolsep= 0.17cm 
\centering
 \setlength{\abovecaptionskip}{ 2 pt}
\caption{Quantitative results of our method and baselines on the S3DIS-Area2 \cite{Dai2017}.}
\label{tab:app_s3dis_area2}
\resizebox{0.95\linewidth}{!}
{
\begin{tabular}{crccc}
\toprule[1.0pt]
& & AP(\%) & AP@50(\%)& AP@25(\%) \\
\toprule[1.0pt]
\multirow{6}{*}{\makecell[l]{Unsupervised}}
& UnScene3D \cite{Rozenberszki2024} &3.2  &6.0  &10.5  \\
& Part2Object \cite{Shi2024} &2.8  &6.2  &8.8  \\
& EFEM \cite{Lei2023} &1.1 &2.9 &9.2 \\
& GrabS-VAE \cite{Zhang2025} &\textbf{6.4}  &\textbf{10.2}  &17.2  \\
& GrabS-Diffusion \cite{Zhang2025} &5.3  &8.1  &13.3  \\
&\textbf{\nickname{} (Ours-VAE)} &5.7  &9.5  &\textbf{20.7} \\
&\textbf{\nickname{} (Ours-Diffusion)} &5.3  &9.9  &16.7 \\
\bottomrule[1.0pt]
\end{tabular}
}
\vspace{-0.2cm}
\end{table}

\begin{table}[th]\tabcolsep= 0.17cm 
\centering
 \setlength{\abovecaptionskip}{ 2 pt}
\caption{Quantitative results of our method and baselines on the S3DIS-Area3 \cite{Dai2017}.}
\label{tab:app_s3dis_area3}
\resizebox{0.95\linewidth}{!}
{
\begin{tabular}{crccc}
\toprule[1.0pt]
& & AP(\%) & AP@50(\%)& AP@25(\%) \\
\toprule[1.0pt]
\multirow{6}{*}{\makecell[l]{Unsupervised}}
& UnScene3D \cite{Rozenberszki2024} &37.6  &58.2  &83.6  \\
& Part2Object \cite{Shi2024} &38.9  &63.5  &81.4  \\
& EFEM \cite{Lei2023} &29.0 &58.3 &64.2 \\
& GrabS-VAE \cite{Zhang2025} &59.5  &78.8  &80.2  \\
& GrabS-Diffusion \cite{Zhang2025} &51.4  &67.0  &67.0  \\
&\textbf{\nickname{} (Ours-VAE)} &66.0  &86.0  &\textbf{91.8} \\
&\textbf{\nickname{} (Ours-Diffusion)} &\textbf{70.0}  &\textbf{91.3}  &91.5 \\
\bottomrule[1.0pt]
\end{tabular}
}
\vspace{-0.2cm}
\end{table}

\begin{table}[th]\tabcolsep= 0.17cm 
\centering
 \setlength{\abovecaptionskip}{ 2 pt}
\caption{Quantitative results of our method and baselines on the S3DIS-Area4 \cite{Dai2017}.}
\label{tab:app_s3dis_area4}
\resizebox{0.95\linewidth}{!}
{
\begin{tabular}{crccc}
\toprule[1.0pt]
& & AP(\%) & AP@50(\%)& AP@25(\%) \\
\toprule[1.0pt]
\multirow{6}{*}{\makecell[l]{Unsupervised}}
& UnScene3D \cite{Rozenberszki2024} &23.9  &49.1  &70.8  \\
& Part2Object \cite{Shi2024} &26.8  &57.4  &75.3  \\
& EFEM \cite{Lei2023} &15.1 &36.5 &49.1 \\
& GrabS-VAE \cite{Zhang2025} &39.2  &66.4  &73.4  \\
& GrabS-Diffusion \cite{Zhang2025} &39.4  &64.8  &68.0  \\
&\textbf{\nickname{} (Ours-VAE)} &41.7  &70.5  &85.6 \\
&\textbf{\nickname{} (Ours-Diffusion)} &\textbf{42.0}  &\textbf{72.3}  &\textbf{89.4} \\
\bottomrule[1.0pt]
\end{tabular}
}
\vspace{-0.2cm}
\end{table}

\begin{table}[th]\tabcolsep= 0.17cm 
\centering
 \setlength{\abovecaptionskip}{ 2 pt}
\caption{Quantitative results of our method and baselines on the S3DIS-Area5 \cite{Dai2017}.}
\label{tab:app_s3dis_area5}
\resizebox{0.95\linewidth}{!}
{
\begin{tabular}{crccc}
\toprule[1.0pt]
& & AP(\%) & AP@50(\%)& AP@25(\%) \\
\toprule[1.0pt]
\multirow{6}{*}{\makecell[l]{Unsupervised}}
& UnScene3D \cite{Rozenberszki2024} &42.6  &63.4  &80.3  \\
& Part2Object \cite{Shi2024} &30.0  &50.5  &76.4  \\
& EFEM \cite{Lei2023} &14.9 &35.7 &45.3 \\
& GrabS-VAE \cite{Zhang2025} &46.4  &66.2  &73.8  \\
& GrabS-Diffusion \cite{Zhang2025} &44.2  &58.0  &62.6  \\
&\textbf{\nickname{} (Ours-VAE)} &55.1  &77.1  &86.2 \\
&\textbf{\nickname{} (Ours-Diffusion)} &\textbf{60.6}  &\textbf{82.8}  &\textbf{92.1} \\
\bottomrule[1.0pt]
\end{tabular}
}
\vspace{-0.2cm}
\end{table}

\begin{table}[th]\tabcolsep= 0.17cm 
\centering
 \setlength{\abovecaptionskip}{ 2 pt}
\caption{Quantitative results of our method and baselines on the S3DIS-Area6 \cite{Dai2017}.}
\label{tab:app_s3dis_area6}
\resizebox{0.95\linewidth}{!}
{
\begin{tabular}{crccc}
\toprule[1.0pt]
& & AP(\%) & AP@50(\%)& AP@25(\%) \\
\toprule[1.0pt]
\multirow{6}{*}{\makecell[l]{Unsupervised}}
& UnScene3D \cite{Rozenberszki2024} &41.3  &70.9  &92.3  \\
& Part2Object \cite{Shi2024} &26.5  &57.4  &83.0  \\
& EFEM \cite{Lei2023} &18.3 &45.2 &53.1 \\
& GrabS-VAE \cite{Zhang2025} &52.1  &80.4  &84.3  \\
& GrabS-Diffusion \cite{Zhang2025} &47.1  &74.5  &76.2  \\
&\textbf{\nickname{} (Ours-VAE)} &\textbf{58.8}  &\textbf{84.4}  &\textbf{93.6} \\
&\textbf{\nickname{} (Ours-Diffusion)} &53.9  &77.2  &91.1 \\
\bottomrule[1.0pt]
\end{tabular}
}
\vspace{-0.2cm}
\end{table}

\subsection{Construction of Multiclass Synthetic Dataset}\label{app:syn_construct}
Similar to GrabS \cite{Zhang2025}, we generate 5000 static scenes that contain 4 to 8 objects. We construct scenes using six object categories from ShapeNet, namely \textit{chair, sofa, telephone, airplane, rifle, cabinet}. Specifically, we generate 4000 and 1000 3D indoor rooms for training and testing, respectively. To prevent data leakage, 3D objects in training scenes are sampled from the ShapeNet validation set, while those in test scenes are selected solely from the ShapeNet test set.

\begin{figure}[t]
\centering 
\centerline{\includegraphics[width=0.3\textwidth]{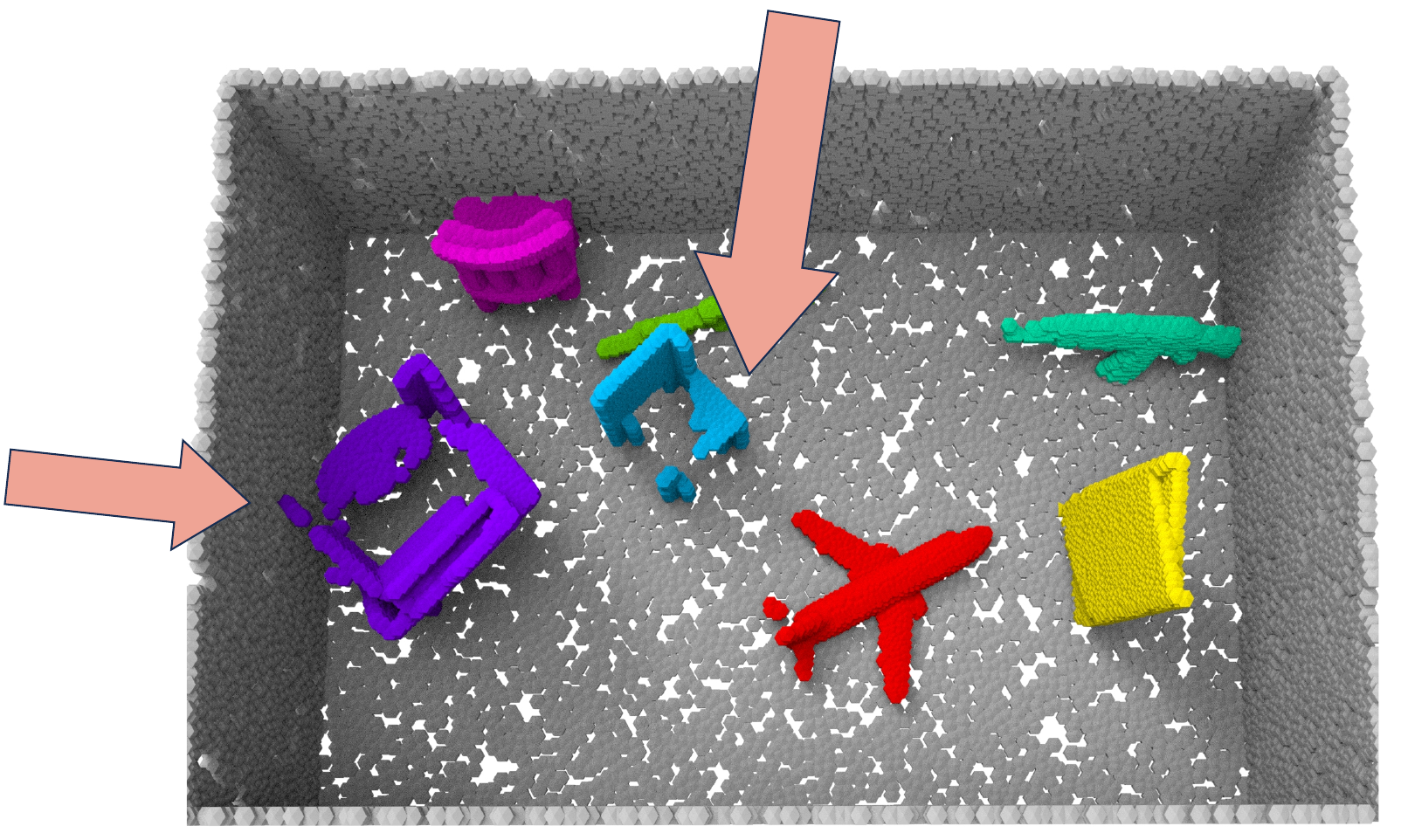}}
\vspace{-0.2cm}
    \caption{Generated Synthetic Dataset}
    \label{fig:sys_occ_app}
    \vspace{-0.5cm}
\end{figure}

\textbf{Occlusion Simulation}: In real-world scenarios, point clouds suffer from incompleteness due to self-occlusion or limitations of scanning angles. For example, when a sofa is placed against a wall on the floor, the scanning device fails to capture the back and bottom surfaces of the sofa. To better simulate real-world scenarios, we generated incomplete point clouds through multi-view projection. Specifically, the camera is positioned at a distance of 2 units from the object, with a randomly sampled pitch angle ranging from 0° to 30° and an azimuth angle from -180° to 180°. The camera FOV is set to 20°, and we randomly generate 12 view configurations. Depth maps are rendered under these configurations and then projected to 3D, obtaining partial object point clouds.
The incomplete object is made by randomly combining 2$\sim$4 partial point clouds. Figure \ref{fig:sys_occ_app} demonstrates the incomplete objects in the generated synthetic dataset.

The aspect ratio of the ground plane in each scene is uniformly sampled between 0.6 and 1.0. All objects share the same size, which is set as 1. To simulate a real scene, we rotate each object around the vertical z-axis from -180° to 180°, then put them on the generated ground. The four sides of the scene are enclosed by generated walls. The resulting point clouds contain only the coordinates without color information. Each scene has 20000 points in total.

\textbf{Object Placement}: To avoid object overlap, we sequentially place objects into the scene and perform collision detection using their bounding boxes against those of the already placed objects. If a collision occurs, the position of objects will be adjusted with a maximum of 1000 attempts. In case no suitable position for the object is found after exceeding 1000 attempts, the scene will be discarded.

\subsection{Training and Evaluation on Synthetic Dataset}\label{app:syn}

\begin{figure*}[th]
\centering 
\centerline{\includegraphics[width=1\textwidth]{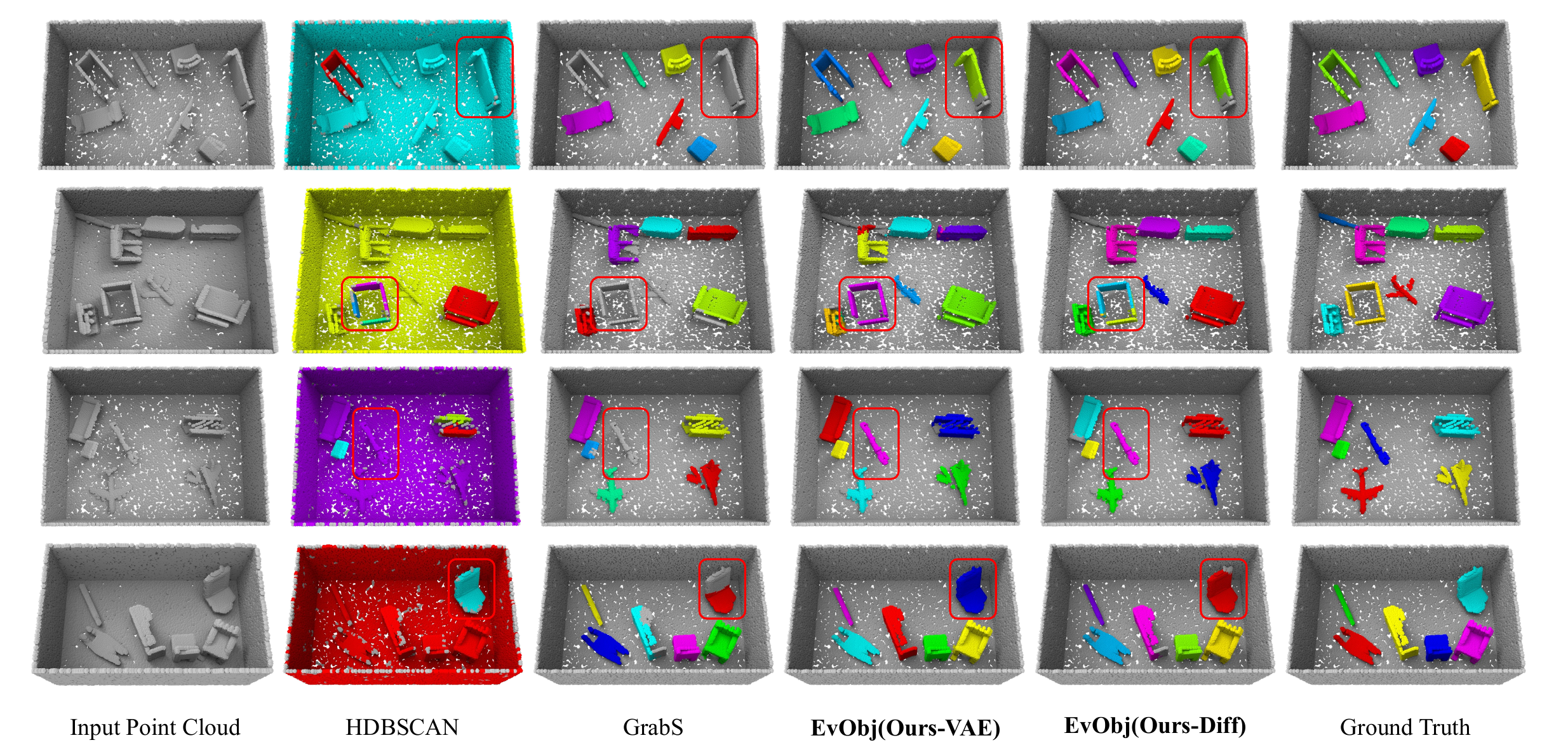}}
\vspace{-0.2cm}
    \caption{More qualitative results on the test set of our Synthetic dataset. Red boxes highlight the differences.}
    \label{fig:synthetic_app}
    \vspace{-0.2cm}
\end{figure*}

\textbf{Training Process}: Training hyperparameters for synthetic dataset are aligned with those of ScanNet. The query number in Mask3D is set as 10 due to the maximum number of objects in this synthetic dataset being 8. We train the model for 200 epochs with a batch size of 10, using the AdamW optimizer configured with a fixed learning rate of 1e-4. 

All object categories \textit{chair, sofa, telephone, airplane, rifle, cabinet} are evaluated on the test set of our synthetic dataset. More visualizations of the results are provided in Figure \ref{fig:synthetic_app}. Benefiting from the object candidate completion module, our model significantly outperforms baseline methods when handling incomplete objects.

\subsection{Training and Evaluation on ScanNet++}
ScanNet++: It is a high-quality 3D indoor scene dataset. Compared with ScanNet, ScanNet++ \cite{yeshwanth2023scannet++} exhibits significantly higher intra-class diversity across object categories. 

To better validate the robustness and generalizability of our method, we conduct experiments under a cross-dataset validation setup. Specifically, we reuse the well-trained segmentation models trained on ScanNet and perform a class-agnostic evaluation on all chairs in the validation set of ScanNet++, which contains 50 3D scenes.

\textbf{Analysis}: Quantitative results are shown in Table \ref{tab:scannetpp}. Our method significantly surpasses all unsupervised baselines on the validation set. The reason is that our method can mitigate the domain gap between synthetic scenes and real scenes and thus better handle intra-class diversity.

\begin{table}[ht]
\tabcolsep= 0.1cm 
\centering
 \setlength{\abovecaptionskip}{ 2 pt}
\caption{Quantitative results of cross dataset validation on vali-
dation set of ScanNet++.}
\label{tab:scannetpp}
\resizebox{1.0\linewidth}{!}{
\begin{tabular}{lrccc}
\toprule[1.0pt]
& & AP(\%) & AP@50(\%)& AP@25(\%) \\
\toprule[1.0pt]
\multirow{4}{*}{\makecell[l]{Unsupervised}} 
& Part2Object \cite{Shi2024} &13.7  &31.8 &57.2 \\
& UnScene3D \cite{Rozenberszki2024} &14.4  &35.6 &65.1 \\
& GrabS \cite{Zhang2025} &19.8  &35.6 &48.1 \\
&\textbf{\nickname{}} &\textbf{42.0}  &\textbf{69.3}  &\textbf{82.7} \\
\bottomrule[1.0pt]
\end{tabular}}
\vspace{-0.3cm}
\end{table}

\subsection{Evaluation on Object Candidate Discerning Module}\label{app:analy}
Table \ref{tab:suff} reports object candidate coverage for our \nickname{} and GrabS, demonstrating that \nickname{} discovers more objects in 3D scenes during RL training, thus yielding superior segmentation performance.

Discovered object candidates are counted as true objects if their IoU with ground truth masks exceeds a threshold. Table \ref{tab:app_suff} presents additional comparative results across different IoU thresholds. After training more than 100 epochs, our \nickname{} consistently outperforms GrabS \cite{Zhang2025} in all cases, while \nickname{} without the discerning module exhibits a significant performance drop, validating the effectiveness of our object candidate discerning module in object discovery.

\begin{table}[h]
\centering
\caption{The sufficiency (\%) of qualified object candidates discovered over epochs on the ScanNet training .}
\label{tab:app_suff}
\resizebox{1\linewidth}{!}{
\begin{tabular}{crcccccc}
\toprule[1.0pt]
&Epoch &100 &200 &300 &400 &500\\
\toprule[1.0pt]
\multirow{3}{*}{\makecell[l]{IoU 50\%}}
&GrabS \cite{Zhang2025} &55.2 &59.4 &61.4 &62.5 &62.6 \\
&\nickname{} without $f_{dis\_evo}$ &65.3 &68.6 &69.6 &69.2 &69.6 \\
&\nickname{} with $f_{dis\_evo}$ &66.3 &70.5 &71.4 &71.7 &71.6 \\
\toprule[1.0pt]
\multirow{3}{*}{\makecell[l]{IoU 60\%}}
&GrabS \cite{Zhang2025} &47.9 &51.3 &52.6 &53.5 &53.7 \\
&\nickname{} without $f_{dis\_evo}$ &52.5 &54.8 &55.3 &55.5 &56.2 \\
&\nickname{} with $f_{dis\_evo}$ &55.1 &59.2 &60.9 &61.2 &61.3 \\
\toprule[1.0pt]
\multirow{3}{*}{\makecell[l]{IoU 70\%}}
&GrabS \cite{Zhang2025} &37.5 &40.0 &40.8 &41.6 &41.7 \\
&\nickname{} without $f_{dis\_evo}$ &38.0 &38.4 &38.6 &38.7 &39.1 \\
&\nickname{} with $f_{dis\_evo}$ &41.3 &45.0 &46.7 &47.5 &46.9 \\
\toprule[1.0pt]
\multirow{3}{*}{\makecell[l]{IoU 80\%}}
&GrabS \cite{Zhang2025} &23.4 &24.5 &24.8 &25.8 &25.8 \\
&\nickname{} without $f_{dis\_evo}$ &21.4 &22.1 &22.0 &21.6 &21.6 \\
&\nickname{} with $f_{dis\_evo}$ &24.8 &29.0 &30.2 &31.4 &30.8 \\
\bottomrule[1.0pt]
\end{tabular}}
\end{table}

\begin{figure*}[t]
\centering 
\centerline{\includegraphics[width=1\textwidth]{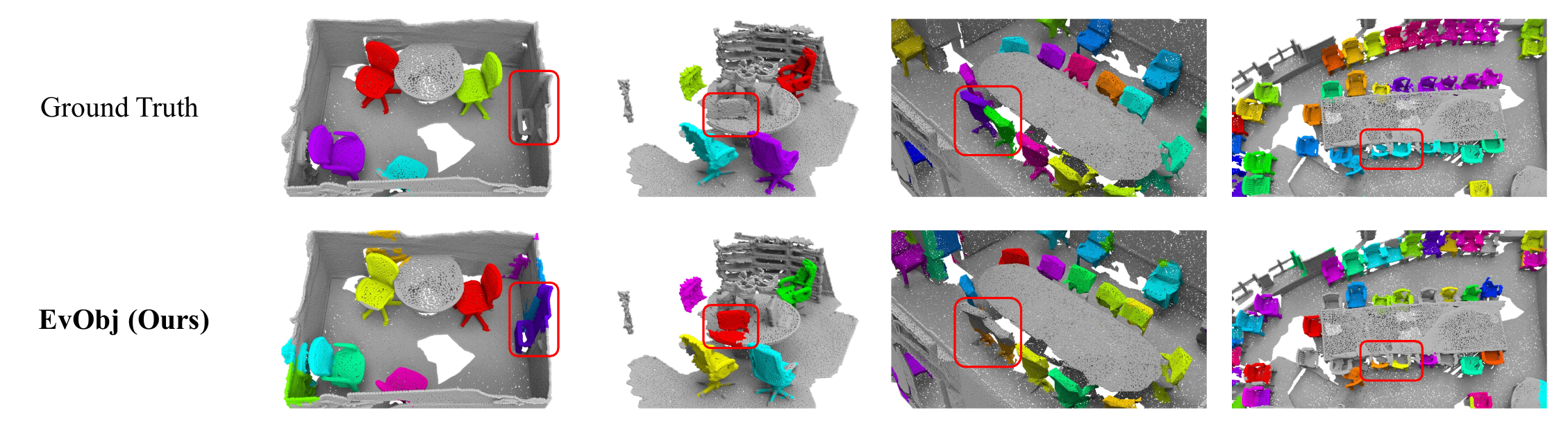}}
\vspace{-0.2cm}
    \caption{Failure cases of our method.}
    \label{fig:failure_case}
    \vspace{-0.2cm}
\end{figure*}

\subsection{Ablation Study on Object Candidate Discerning Module}
For training the discerning module, we adopt SparseUNet \cite{Graham2018} as the backbone network and further conduct an ablation study to investigate the impact of backbone selection for this module. Specifically, we evaluate two alternative backbones: PointNet++ \cite{Qi2017}, and Point Transformer \cite{zhao2021point}, which yield similar performance as shown in Table \ref{tab:ablative_appendix}.

\begin{table}\tabcolsep= 0.12cm 
\centering
 \setlength{\abovecaptionskip}{ 2 pt}
\caption{Ablation results on the validation set of ScanNet.}
\label{tab:ablative_appendix}
\resizebox{1.0\linewidth}{!}{
\begin{tabular}{lccc}
\toprule[1.0pt]
 &AP(\%) &AP@50(\%) &AP@25(\%)\\
\toprule[1.0pt]
\textbf{Ablations on Discerning Module}\\
\textbf{\nickname{}$_{SparseUNet\cite{Graham2018}}$}  &\textbf{55.0} &\textbf{76.9}  &\textbf{88.2} \\
\nickname{}$_{PointNet++\cite{Qi2017}}$ &54.5  &76.5  &87.4 \\
\nickname{}$_{PointTransformer \cite{zhao2021point}}$ &54.7 &76.5 &87.8 \\
\toprule[1.0pt]
\textbf{Ablations on Chamfer Distance $\delta_c$}\\
$\delta_c$ = 0.08 &48.0 &67.0  &78.9 \\
$\delta_c$ = 0.09 &53.2 &74.9  &86.0 \\
$\delta_c$ = 0.10 &\textbf{55.0} &\textbf{76.9}  &\textbf{88.2} \\
$\delta_c$ = 0.11 &52.5 &75.6  &88.1 \\
$\delta_c$ = 0.12 &51.0 &73.8  &87.7 \\
\toprule[1.0pt]
\end{tabular}
}\vspace{-0.5cm}
\end{table}

\subsection{Ablation Study on Chamfer Distance}
During reinforcement learning, the reward is defined by the reconstruction error of the object-centric network, which is measured by the Chamfer Distance $\delta_c$ between the input and output point clouds of the object-centric network. Given the critical role of rewards in guiding the agent, we perform five ablation studies on the threshold $\delta_c$. As shown in Table \ref{tab:ablative_appendix}, the results are robust to the choice of this hyperparameter, and we set $\delta_c$ as 0.10 in our main experiments.

\subsection{Time and Memory Costs}
Training the entire pipeline of \nickname{} takes 104 hours. Specifically, training the discerning module takes 0.5 hours with 4 GB GPU memory, the completion module takes 6 hours with 19 GB, the object-centric net takes 29 hours with 8 GB, and the policy network takes 69 hours with 20 GB.

While the total training time of our method is slightly longer than the baseline GrabS \cite{Zhang2025}, which takes 91 hours, our approach maintains the same inference speed. On average, it processes one scene in 0.063 seconds with 5 GB of GPU memory. All experiments are conducted on a single NVIDIA RTX 3090 GPU with an AMD R9 5900X CPU.

\subsection{Plug-and-Play Mask Refinement}
To validate the generalizability of our method, we apply our discerning module as a post-processing (plug-and-play) technique to UnScene3D \cite{Rozenberszki2024} and Part2Object \cite{Shi2024}, refining their predicted masks. We conduct experiments on the chair category of the validation set of ScanNet, treating all predicted masks as chairs. Table \ref{tab:plug_and_play} shows that both baselines can be significantly improved, highlighting the versatility of our method.

\begin{table}\tabcolsep= 0.12cm 
\centering
 \setlength{\abovecaptionskip}{ 2 pt}
\caption{Quantitative results of Plug and Play on the validation set of ScanNet.}
\label{tab:plug_and_play}
\resizebox{1.0\linewidth}{!}{
\begin{tabular}{lccc}
\toprule[1.0pt]
 &AP(\%) &AP@50(\%) &AP@25(\%)\\
\toprule[1.0pt]

UnScene3D \cite{Rozenberszki2024} &37.2 &62.4 &79.2 \\
UnScene3D$_{f_{dis\_evo}}$ &\textbf{41.1} &\textbf{65.7} &\textbf{79.4} \\
\toprule[1.0pt]
Part2Object \cite{Shi2024}  &34.4 &56.8 &73.9\\
Part2Object$_{f_{dis\_evo}}$  &\textbf{39.2} &\textbf{64.5} &\textbf{81.8} \\

\toprule[1.0pt]
\end{tabular}
}\vspace{-0.5cm}
\end{table}

\subsection{Visualizations of Failure Cases}
Our method \nickname{} presents two types of failure cases as shown in Figure \ref{fig:failure_case}. For the first type, it yields erroneous segmentation for objects with similar shape to the target objects. For instance, it incorrectly segments parts of walls and a trash bin as a single chair. The second type involves segmenting only object parts rather than whole instances. For example, it only segments the legs of the chair instead of the entire chair. We attribute this issue to the completion module.

\begin{figure*}[t]
\centering 
\centerline{\includegraphics[width=1\textwidth]{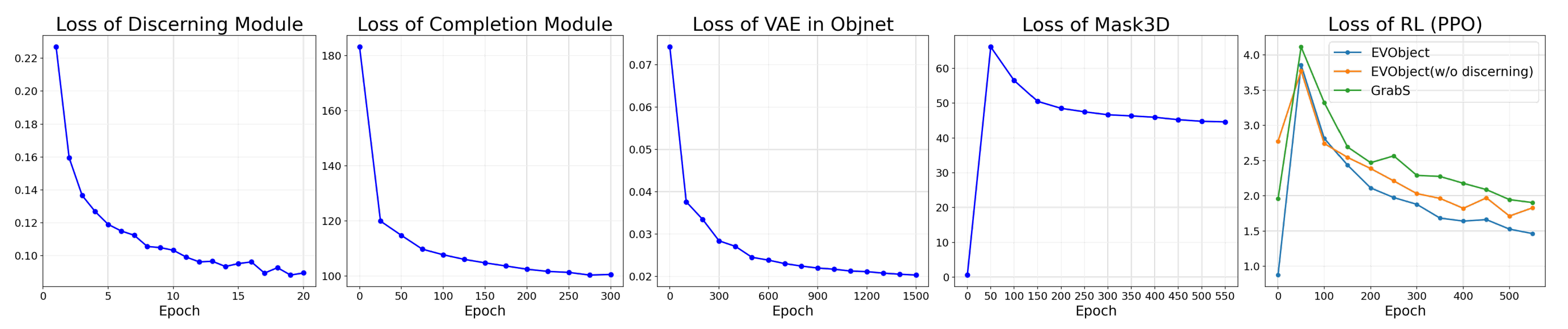}}
\vspace{-0.2cm}
    \caption{Training losses of all modules in our pipeline.}
    \label{fig:appendix_loss}
    \vspace{-0.2cm}
\end{figure*}

\subsection{Pipeline Complexity and Error Accumulation}

As we introduce two additional modules, namely the discerning module and completion module, the complexity of the overall pipeline is increased. We analyze whether this introduces difficulties in training and leads to error accumulation between modules.

Figure \ref{fig:appendix_loss} shows the training losses for the discerning, completion, object-centric, segmentation, and policy modules, highlighting the stable convergence and ease of training achieved by our modular pipeline.

We analyze error accumulation by applying the trained agent on training scenes in ScanNet and then obtaining object masks from the discerning, completion, and object-centric networks. Specifically, for the discerning module, we compute the IoU between the predicted masks and the Ground Truth masks. 
For the completion module, we compute the point distances from scene point clouds to the completed object point clouds, and filter them by a threshold of 0.1, and finally get masks from the completion module to compute IoU with Ground Truth. For object-centric networks, we also compute the point distance from scene points to the generated mesh, then get masks by filtering with a 0.1 threshold. This follows the same procedure as the completion module for matching and IoU calculation.

The average IoUs against ground truth are 42.0, 50.7, and 46.9, respectively, indicating that error accumulation is not significant.

\end{document}